\documentclass[11pt,a4paper]{article}

\usepackage[utf8]{inputenc}
\usepackage[T1]{fontenc}
\usepackage{lmodern}
\usepackage[english]{babel}
\usepackage[margin=2.5cm]{geometry}
\usepackage{graphicx}
\usepackage{booktabs}
\usepackage{natbib}
\usepackage{amsmath}
\usepackage{amssymb}
\usepackage{setspace}
\usepackage{caption}

\usepackage{amsthm}
\usepackage{enumitem}
\usepackage{xcolor}
\usepackage{microtype}

\usepackage[colorlinks=true, citecolor=blue, linkcolor=blue, urlcolor=blue]{hyperref}
 \newtheorem{theorem}{Theorem}[subsection]

\newtheorem{definition}[theorem]{Definition}

\newtheorem{assumption}{Assumption}
\setcitestyle{round}

\let\oldfootnote\footnote
\renewcommand{\footnote}{\fontsize{9}{11}\selectfont\oldfootnote}

\newcommand{\bx}{\boldsymbol{x}}
\newcommand{\bX}{\boldsymbol{X}}
\newcommand{\bY}{\boldsymbol{Y}}
\newcommand{\bZ}{\boldsymbol{Z}}
\newcommand{\bbeta}{\boldsymbol{\beta}}
\newcommand{\btheta}{\boldsymbol{\theta}}
\newcommand{\bLambda}{\boldsymbol{\Lambda}}
\newcommand{\bEta}{\boldsymbol{\eta}}
\newcommand{\bA}{\boldsymbol{A}}
\newcommand{\bD}{\boldsymbol{D}}
\newcommand{\bW}{\boldsymbol{W}}

\newcommand{\cH}{\mathcal{H}}
\newcommand{\cX}{\mathcal{X}}
\newcommand{\cL}{\mathcal{L}}
\newcommand{\cR}{\mathcal{R}}
\newcommand{\KK}{\mathbb{K}}
\newcommand{\PP}{\mathbb{P}}
\newcommand{\RR}{\mathbb{R}}
\newcommand{\NN}{\mathbb{N}}

\newenvironment{revision}{}{}
\DeclareRobustCommand{\revtext}[1]{#1}
\DeclareRobustCommand{\revheading}[1]{#1}
  
\title{Towards a Statistical Understanding of Neural Networks: Beyond the Neural Tangent Kernel Theories}

\author{
  Yicheng Li\textsuperscript{1,2},
  Haobo Zhang\textsuperscript{2},
  Jianfa Lai\textsuperscript{2},
  Qian Lin\textsuperscript{2}, and
  Jun S. Liu\textsuperscript{2}\thanks{Corresponding author: \texttt{junsliu@tsinghua.edu.cn}}\\[0.5em]
  \textsuperscript{1}KLATASDS-MOE, School of Statistics,\\
  East China Normal University, Shanghai, China\\
  \textsuperscript{2}Department of Statistics and Data Science,\\
  Tsinghua University, Beijing, China\\[0.35em]
  \texttt{ycli@sfs.ecnu.edu.cn}\\
  \texttt{zhang-hb21@mails.tsinghua.edu.cn}\\
  \texttt{jianfalai@mail.tsinghua.edu.cn}\\
  \texttt{qianlin@tsinghua.edu.cn}\\
  \texttt{junsliu@tsinghua.edu.cn}
}
\date{}

\hypersetup{
  pdftitle={Towards a Statistical Understanding of Neural Networks: Beyond the Neural Tangent Kernel Theories},
  pdfauthor={Yicheng Li, Haobo Zhang, Jianfa Lai, Qian Lin, and Jun S. Liu},
  pdfkeywords={neural networks, generalization ability, feature learning, kernel regression, over-parameterized Gaussian sequence model},
}

\begin{document}
\maketitle

\begin{abstract}
  A primary advantage of neural networks lies in their feature learning characteristics, which is challenging to theoretically analyze due to the complexity of their training dynamics.
We examine feature learning and its potential benefits for generalization from a statistical perspective.
After reviewing the neural tangent kernel (NTK) theory and recent results in kernel regression, which address the generalization issue of sufficiently wide neural networks, we examine limitations and implications of the fixed kernel theory (as the NTK theory) and review recent theoretical advancements in feature learning.
Moving beyond theories with fixed features, we consider neural networks as adaptive feature models.
Finally, we propose an over-parameterized Gaussian sequence model as a prototype for the adaptive feature model to study feature learning characteristics and motivate their future analysis for neural networks.
 \end{abstract}



\section{Introduction}

In recent years, deep neural networks have made remarkable accomplishments in many application areas, whereas their theoretical understanding has lagged far behind.
Although neural network modeling has a long history, recent developments in data availability, computing resources, and network architecture designs were believed to be the key to their somewhat mysteriously outstanding performance.

Numerous topics have been explored regarding neural network theory, such as those discussed in recent review papers and references therein \citep{fan2021selective,bartlett2021deep,belkin2021fit,suh2024survey}.
In this paper, we consider the generalization ability of neural networks within the nonparametric regression framework.
\revtext{Let a compact set $\cX \subset \RR^{d}$ be the input space, let $ \mathcal{Y} \subseteq \RR$ be the output space, and let $\mu$ be the input distribution on $\cX$.}
Suppose that the $n$ i.i.d. draws $\{ (\bx_{i}, y_{i}) \}_{i=1}^{n}$ are  sampled from the model
\begin{equation}\label{main data model}
    y = f^{*}(\bx) + \epsilon,  \quad \bx \sim \mu,
\end{equation}
\revtext{where $ f^{*} \in L^{2}(\cX,\mu)$ is the true function, and the noise $\epsilon$, independent of $\bx$, has zero mean and variance $\sigma_{0}^{2} > 0$.}
Throughout the paper, we denote $\bX = (\bx_{1}, \ldots, \bx_{n}) \in \RR^{d \times n}$, and $\bY = (y_{1}, \ldots, y_{n})^{\top}$.
For any estimator $\hat{f}$ (which can be generated by a neural network, a kernel regression model, etc.), we will investigate the $L^{2}$-norm generalization error (or excess risk) given by
\begin{equation}\label{eq l2 gen error}
\begin{aligned}
    \cR(\hat{f}) = \left\| \hat{f} - f^{*}\right\|^2_2
\stackrel{\Delta}{=} \mathbb{E}_{\bx \sim \mu} \left[ \left( \hat{f}(\bx) - f^{*}(\bx) \right)^{2} \right].
\end{aligned}
\end{equation}

It is natural to ask the following questions about neural networks:
\begin{itemize}
\item [(i)] How can the generalization ability of neural networks be characterized from a statistical viewpoint?
For instance, can we establish convergence rates of the generalization error with respect to sample sizes?
Do we know the minimax optimality of a neural network method?
\item [(ii)] Why do neural networks outperform other existing methods in many applications?
Can we understand when and how?
\end{itemize}

To answer the first question,  the line of work of \cite{bauer2019deep}, \cite{schmidt2020nonparametric}, and \cite{suzuki2018adaptivity}  considered the ``algorithm-independent controls'' of the generalization error, a name coined by \cite{fan2021selective}.
By carefully selecting the candidate function class represented by a specific neural network architecture, they obtained generalization error bounds for the empirical risk minimization estimator over the chosen function class.
\begin{revision}
However, the empirical risk minimization estimator is not necessarily the output of a specified training procedure.
For a parameterized neural network $f_{\btheta}$, training acts on the empirical risk
\begin{equation}\label{eq empirical risk neural network}
    \cL_n(\btheta) = \frac{1}{n}\sum_{i=1}^{n}\left(y_i-f_{\btheta}(\bx_i)\right)^2.
\end{equation}
The map $\btheta\mapsto\cL_n(\btheta)$ is generally nonconvex.
Thus, an estimator-level guarantee does not by itself show that a specified training procedure reaches a parameter producing that estimator.
Initialization and training choices can lead to different parameter trajectories and predictors.
\end{revision}

For comparison, our current paper focuses on {\it algorithm-dependent controls} \citep{fan2021selective}, which take into account the training dynamics of neural networks.
In practice, training dynamics are usually difficult to understand and can lead to drastically different generalization behaviors.
This makes the theoretical analysis of training dynamics one of the most intriguing and challenging tasks.

Our starting point is the seminal \textit{neural tangent kernel} (NTK) theory \citep{jacot2018neural,arora2019exact,lee2019wide}, which establishes a connection between the training dynamics of sufficiently wide neural networks and the kernel gradient flow (Section \ref{section ntk theory}).
Due to the success of the NTK theory, we have experienced a renaissance of kernel regression.
We will review some important topics about the generalization ability of kernel regression, addressing both fixed-dimensional (Section \ref{subsection fixed d kr}) and high-dimensional settings (Section \ref{subsection large d kr}).

The answer to the second question is even more complicated.
It is believed that the superiority of neural networks stems from their feature learning characteristics, which cannot be addressed by the NTK theory.
\revtext{Recent work has examined finite-width NTK behavior, the scope and limitations of NTK theory, and the evolution of NTK eigenvectors at the edge of stability \citep{seleznova2022finite,golikov2022ntk,jiang2025evolution}.}
Therefore, another theme of this paper is to go beyond the NTK theory and study feature learning characteristics of neural networks.
Specifically, in Section \ref{section from fix to feature}, we discuss limitations and implications of the NTK theory (as the fixed kernel regression) and review some recent advances in feature learning theory.
In Section \ref{section over-para gsm}, we consider an \textit{adaptive feature model} to emulate the feature learning process of neural networks.
\revtext{Furthermore, we introduce the \textit{over-parameterized Gaussian sequence model} as a viable prototype for the adaptive feature model.}
\revtext{\citet{lecue2025sharp} formalize the distinction between alignment for a fixed representation and feature learning, whereas \citet{gavrilopoulos2024geometrical} train a representation in the first layer and then analyze kernel ridge regression with the resulting conjugate kernel learned from the data, including an excess risk result for a single neuron example.}
\revtext{Relative to these works, our contribution is the Gaussian sequence prototype with learnable rotations and rescalings, rather than a new general definition of feature learning, a training mechanism for neural networks, or a mathematical guarantee.}
\revtext{This model offers a tractable way to formulate questions about training dynamics and feature learning.}
Finally, we discuss some future questions in Section \ref{section discussion}.
We hope that our review and the proposed prototype can offer new insights and directions for developing theoretical understandings of neural networks.

{\bf Notations}: The asymptotic notations, $O(\cdot),$ $o(\cdot),$ $\Omega(\cdot)$ and $\Theta(\cdot)$, will be used throughout the paper: We say $a_{n} = \Omega(b_{n})$ if $ b_{n} = O(a_{n})$; $a_{n} = \Theta(b_{n})$ if $ a_{n} = O(b_{n})$ and $a_{n} = \Omega(b_{n})$.
We also write $a_n \asymp b_n$ for $a_n = \Theta(b_n)$, and $a_n \lesssim b_n$ for $a_n = O(b_n)$.
The corresponding probabilistic versions of the asymptotic notations will also be used: $O_{\PP}(\cdot), o_{\PP}(\cdot), \Omega_{\PP}(\cdot)$ and $ \Theta_{\PP}(\cdot)$.
For example, we say that random variables $ X_{n}, Y_{n} $ satisfy $ X_{n} =  O_{\PP}(Y_{n}) $ if $\forall \ \varepsilon > 0$, there exist constants $C_{\varepsilon} $ and $ N_{\varepsilon}$ such that $ \PP\left( |X_{n}| \ge C_{\varepsilon} |Y_{n}| \right) \le \varepsilon$ for $n > N_{\varepsilon}$.
For $\alpha>0$, we write $\lfloor \alpha \rfloor$ as the largest integer not exceeding $\alpha$.
We use boldface letters to denote column vectors and matrices;  $\| \bx \|_{2}$ is the $ \ell^{2}$-norm of vector $\bx$.
 \section{\revheading{Neural tangent kernel theory and kernel regression}}\label{section kernel regression}

\subsection{Neural tangent kernel theory}\label{section ntk theory}
The neural tangent kernel (NTK) theory \citep{jacot2018neural,arora2019exact,lee2019wide} is one of the most successful tools for analyzing neural networks' training dynamics and generalization ability.
A major challenge in studying neural networks is the highly non-convex nature of the target function.
The basic idea of the NTK theory is that when a neural network's width is sufficiently large, the network's training dynamics can be approximated well by the kernel gradient flow for the corresponding neural tangent kernel, which is a convex problem.
Numerous papers discuss various topics within the NTK theory in different settings, e.g. \cite{lai2023generalization}, \cite{du2019gradient}, \cite{huang2020deep}, \cite{arora2019fine}, \cite{li2024eigenvalue}, and \cite{nitanda2020optimal}.

A purpose of our introduction of the NTK theory is to emphasize the importance of studying kernel regression, which prepares us for proposing the adaptive feature approach in subsequent sections.
Therefore, here we only review one result from \cite{lai2023generalization} to establish the connection between neural networks and kernel gradient flow.
Specifically, \cite{lai2023generalization} considered the following two-layer fully connected ReLU neural network with width $m$:
\begin{equation}\label{eq relu two bias}
    f_{\btheta}^{m}(\bx)=\frac{1}{\sqrt{m}} \sum_{r=1}^m a_r \sigma\left( \boldsymbol{w}_{r}^{\top} \bx + b_{r}\right) + b,
\end{equation}
where $ \btheta = \text{vec}(\{ a_{r}, \boldsymbol{w}_{r}, b_{r}, b, r=1,\ldots,m \})$ and $\sigma(\cdot)$ is the ReLU function.
The neural network is trained through the gradient flow of parameters $\btheta(t)$ to minimize the mean-squared loss:
\begin{displaymath}
    \cL\left(f_{\btheta}^m\right)=\frac{1}{2 n} \sum_{i=1}^n\left(y_i-f_{\btheta}^m\left(\bx_i\right)\right)^2,
\end{displaymath}
i.e.,
\begin{displaymath}
    \frac{\mathrm{d}}{\mathrm{d} t} \btheta(t) = -\nabla_{\btheta} \cL\left(f_{\btheta(t)}^m\right).
\end{displaymath}
Then, $f_{\btheta(t)}^m$ is the neural network estimator at time $t$ (see \citealt{lai2023generalization}, Section 3 for more details).
\begin{revision}
The first-order linearization of $f_{\btheta}^{m}$ around $\btheta(0)$ has tangent features $\nabla_{\btheta} f_{\btheta(0)}^{m}(\bx)$, whose inner products define the corresponding NTK:
\begin{displaymath}
    k_{\mathrm{NTK}}(\bx,\bx') = \left\langle \nabla_{\btheta} f_{\btheta(0)}^{m}(\bx), \nabla_{\btheta} f_{\btheta(0)}^{m}(\bx') \right\rangle.
\end{displaymath}
Accordingly, denote $\hat{f}_{t}^{\mathrm{NTK}}$ as the kernel gradient flow estimator (defined later in Definition \ref{def of kgf}) with this kernel.
\end{revision}
Under certain initialization, \cite{lai2023generalization}, Proposition 2, showed that for any $\varepsilon > 0$, if the width $m$ is sufficiently large, then
\begin{displaymath}
    \sup _{t \geq 0} \sup _{\bx \in \cX}\left|f_{\btheta(t)}^m(\bx) - \hat{f}_t^{\mathrm{NTK}}(\bx)\right| \leq \varepsilon,
\end{displaymath}
and
\begin{displaymath}
\sup _{t \geq 0}\left|\cR\left(f_{\btheta(t)}^m\right)-\cR\left(\hat{f}_t^{\mathrm{NTK}}\right)\right| \leq \varepsilon,
\end{displaymath}
hold with probability at least $1 - o_{m}(1)$, where the randomness comes from the initialization of the parameters.
In other words, as the width $m$ tends to infinity, the neural network estimator $ f_{\btheta(t)}^m $ uniformly converges to $ \hat{f}_t^{\mathrm{NTK}} $, and the corresponding generalization error $ \cR(f_{\btheta(t)}^m )$ is well approximated by $\cR(\hat{f}_t^{\mathrm{NTK}}) $.
\revtext{Consequently, the generalization ability of the neural network estimator can be studied through the kernel gradient flow estimator.}
Similar results have also been derived for other neural network architectures \citep[etc.]{arora2019exact,tirer2022kernel}, making the kernel gradient flow a reasonable alternative for sufficiently wide neural networks and leading to a resurgence in the study of kernel regression.

\subsection{Preliminaries of kernel regression}\label{subsection preli of kr}

In kernel regression (or kernel method), we are given a pre-specified kernel function $k(\cdot, \cdot): \cX \times \cX \to \RR $, which is supposed to be positive-definite, symmetric, and continuous.
Without loss of generality, we assume that the kernel function is bounded by 1, that is, $ \sup_{\bx \in \cX} k(\bx,\bx) \le 1.$ Then, there exists a corresponding function space $\cH \subset L^{2}(\cX,\mu)$, called the reproducing kernel Hilbert space (RKHS), with inner product $\langle \cdot, \cdot \rangle_{\cH}$ and norm $ \| \cdot \|_{\cH}$.
Mercer's theorem (see \citealt{steinwart2008support}, Theorem 4.49) shows that there exists a non-increasing summable sequence $\{ \lambda_{j} \}_{j=1}^{\infty} \subset (0,\infty)$ and a family of functions $\{ \psi_{j} \}_{j=1}^{\infty} \subset \cH $, such that
\begin{equation}\label{eq mercer decom}
    k\left(\bx, \bx^{\prime}\right)=\sum_{j = 1}^{\infty} \lambda_j \psi_{j}(\bx) \psi_{j}\left(\bx^{\prime}\right), \quad \bx, \bx^{\prime} \in \cX,
\end{equation}
where the convergence is absolute and uniform. $\{ \psi_{j} \}_{j=1}^{\infty}$ and $\{ \lambda_{j} \}_{j=1}^{\infty}$ are the eigenfunctions and eigenvalues, respectively, of the kernel function $k$ and the RKHS $\cH$.
The set of functions $ \{ \psi_{j} \}_{j=1}^{\infty} $ can be assumed to be an orthonormal basis of $L^{2}(\cX,\mu)$ without loss of generality.
The RKHS $ \cH $ can be expressed as
\begin{displaymath}
  \cH = \left\{\sum_{j=1}^{\infty} a_j \lambda_j^{\frac{1}{2}} \psi_{j}(\cdot): ~\left(a_j\right)_{j=1}^{\infty} \in \ell^{2} \right\},
\end{displaymath}
\begin{revision}
where $\ell^{2}$ denotes the space of square-summable sequences.
The corresponding norm is
\end{revision}
\begin{displaymath}
    \left\| \sum\limits_{j=1}^{\infty} a_j \lambda_j^{\frac{1}{2}} \psi_{j}(\cdot) \right\|_{\cH} = \left(\sum\limits_{j=1}^{\infty} a_j^{2} \right)^{\frac{1}{2} } .
\end{displaymath}
For more details on RKHS, see \cite{steinwart2008support} and \cite{steinwart2012_MercerTheorem}.

The basic idea of kernel regression is to estimate $f^{*}$ using candidate functions from $\cH$ \citep{Cucker2001OnTM,Kohler2001NonparametricRE,steinwart2008support}.
A large class of kernel regression estimators is collectively introduced as spectral algorithms \citep{rosasco2005_SpectralMethods,caponnetto2006optimal,gerfo2008_SpectralAlgorithms}.
The two most widely studied are the  \textit{kernel ridge regression} and \textit{kernel gradient flow}.
For a kernel function $k$, we denote
\begin{displaymath}
  \KK(\bX, \bX)=\left(k\left(\bx_i, \bx_j\right)\right)_{n \times n}, ~~~~ \KK(\bx, \bX)=\left(k\left(\bx, \bx_1\right), \ldots, k\left(\bx, \bx_n\right)\right).
\end{displaymath}

\begin{definition}[Kernel ridge regression, KRR]\label{def of krr}
For a given kernel function $k$, corresponding RKHS $\cH$ and any $0<\lambda<\infty$, kernel ridge regression constructs an estimator $\hat{f}_{\lambda}$ by solving the penalized least square problem
\begin{displaymath}
    \hat{f}_\lambda = \underset{f \in \cH}{\arg \min } \left(\frac{1}{n} \sum_{i=1}^n\left(y_i-f\left( \bx_{i} \right)\right)^2+\lambda\|f\|_{\cH}^2\right),
\end{displaymath}
where $ \lambda $ is the regularization parameter.
The explicit expression of $\hat{f}_\lambda$ is
\begin{displaymath}
    \hat{f}_{\lambda}(\bx) = \KK(\bx,\bX) \left( \KK(\bX,\bX) + n \lambda \mathbf{I}_{n} \right)^{-1} \bY.
\end{displaymath}
\end{definition}

\begin{definition}[Kernel gradient flow, KGF]\label{def of kgf}
For a given kernel function $k$, corresponding RKHS $\cH$, and any $0<t<\infty$, kernel gradient flow constructs an estimator $\hat{f}_{t}$ by solving the differential equation
\begin{equation}\label{def dif eq of kgf}
        \frac{\mathrm{d}}{\mathrm{d}t} \hat{f}_{t}(\bx) = -\frac{1}{n} \KK(\bx,\bX) \left( \hat{f}_{t}(\bX) - \bY \right),
\end{equation}
and let $\hat{f}_{0}(\bx) \equiv 0$.
For any $t>0$, when the matrix $\KK(\bX,\bX)$ is strictly positive definite, the explicit expression of $\hat{f}_{t}$ is
\begin{equation}\label{eq explicit expression}
    \hat{f}_{t}(\bx) = \KK(\bx,\bX)  \KK(\bX,\bX)^{-1} \left( \mathbf{I}_{n} - e^{-\frac{1}{n} \KK(\bX,\bX) t  } \right) \bY.
\end{equation}

\end{definition}

\begin{revision}
For finite $t$, the differential equation \eqref{def dif eq of kgf} defines the kernel gradient flow estimator $\hat{f}_{t} \in \cH$ minimizing the training loss $ \| \hat{f}_{t}(\bX) - \bY \|_{2}^{2} / n$.
When $\KK(\bX,\bX)$ is strictly positive definite, only its limit as $t \to \infty$ yields the \textit{kernel interpolation estimator}, which is the $\lambda \to 0$ limit of kernel ridge regression:
\begin{equation}\label{eq KI estimator}
    \hat{f}_{\mathrm{inter}}(\bx) = \KK(\bx,\bX)  \KK(\bX,\bX)^{-1} \bY,
\end{equation}
which will be of independent interest.

In the rate comparisons considered below, $t$ plays the same role as the regularization parameter $\lambda$ in KRR: when $t \asymp \lambda^{-1}$, their generalization rates coincide up to the saturation effect of KRR, which will be introduced later.
\end{revision}
In the definition of general spectral algorithms, both $t$ and $\lambda$ can be unified as a regularization parameter in the spectral algorithm's filter function (see, e.g., \citealt{zhang2024optimality}, Definition 1).
Additionally, the kernel gradient flow is a continuous version of the kernel gradient descent, which has similar theoretical properties and is used more frequently in practice.

In the following, we focus on kernel ridge regression and kernel gradient flow.
For studies of general spectral algorithms, see \cite{lin2018_OptimalRates}, \cite{blanchard2018_OptimalRates}, \cite{lin2020_OptimalConvergence}, \cite{zhang2024optimality}, and \cite{li2024generalization}.

\subsection{Kernel regression in fixed dimensions}\label{subsection fixed d kr}

\subsubsection{Assumptions}
In fixed dimensions, we disregard the constants' dependence on the input dimension $d$.
The first commonly used assumption is the eigenvalue decay rate of the eigenvalues $\{\lambda_{j}\}_{j=1}^{\infty}$.
\begin{assumption}[Eigenvalue decay rate, EDR]\label{ass EDR}
Suppose that the eigenvalue decay rate (EDR) of $\cH$ is $\beta > 1$.
That is, there exist positive constants $c$ and $C$ such that
 \begin{displaymath}
   c j^{- \beta} \le \lambda_{j} \le C j^{-\beta}, \quad  \forall j=1,2,\ldots.
 \end{displaymath}
\end{assumption}
Assumption \ref{ass EDR} holds for many kernels, e.g., the Laplacian kernel, Matérn kernel, neural tangent kernel, etc., and is also closely related to the effective dimension or the capacity condition of RKHS \citep{Caponnetto2007OptimalRF}.

The second widely adopted assumption is the source condition, which characterizes the relative smoothness of $f^{*}$ with respect to $\cH$.
To introduce this, we need the following definition of the \textit{interpolation space} (or power space) $[\cH]^s $ for any $s \ge 0$, which is defined as
\begin{equation}\label{def of Hs}
  [\cH]^s = \left\{\sum_{j =1}^{\infty} a_j \lambda_j^{\frac{s}{2}} \psi_{j}(\cdot): \left(a_j\right)_{j =1}^{\infty} \in \ell^{2} \right\},
\end{equation}
equipped with the norm
\begin{displaymath}
    \left\| \sum\limits_{j=1}^{\infty} a_j \lambda_j^{\frac{s}{2}} \psi_{j}(\cdot) \right\|_{[\cH]^s} = \left(\sum\limits_{j=1}^{\infty} a_j^{2} \right)^{\frac{1}{2} } .
\end{displaymath}
Specifically, we have $[\cH]^1 = \cH$.
\revtext{For $0 < s_{1} < s_{2}$, the inclusion maps $ [\cH]^{s_{2}} \hookrightarrow[\cH]^{s_{1}} \hookrightarrow[\cH]^0 $ are compact \citep{fischer2020_SobolevNorm}; here $\hookrightarrow$ denotes a continuous embedding.}
The functions in $[\cH]^{s}$ with smaller $s$ are less ``smooth'', which are harder for a kernel regression algorithm to estimate.
In fact, another equivalent definition of $[\cH]^{s}$ is through the \textit{real interpolation} in functional analysis \citep{tartar2007introduction,sawano2018theory,steinwart2012_MercerTheorem}.
\begin{assumption}[Source condition]\label{ass source condition}
Suppose that for some $s \ge 0 $, there is a constant $R > 0 $ such that $f^{*} \in [\cH]^{s}$ and
  \begin{displaymath}
    \| f^{*} \|_{[\cH]^{s}} \le R.
  \end{displaymath}
We refer to $s$ as the source condition of $f^{*}$.
\end{assumption}

\subsubsection{Learning curve results}\label{subsection lc fix d}

The first question of interest is the minimax optimality of kernel regression.
In the framework of eigenvalue decay rate and source conditions, the minimax optimality was first established in the well-specified case ($f^{*} \in \cH$, or the source condition $s \ge 1$) in \cite{caponnetto2006optimal,Caponnetto2007OptimalRF}.
Then, extensive subsequent literature \citep{steinwart2009_OptimalRates,fischer2020_SobolevNorm,zhang2023optimality_2,PillaudVivien2018StatisticalOO,Celisse2020AnalyzingTD,dieuleveut2016nonparametric,JMLR:v23:21-0570,li2023asymptotic} studied  mis-specified case (source condition $0<s<1$).
Among them, \cite{fischer2020_SobolevNorm} firstly considered the embedding property of $\cH$: we say that $\cH$ has an embedding property of order $\alpha \in (0,1]$ if $[\cH]^\alpha$ can be continuously embedded into $L^\infty(\cX,\mu)$, i.e., the operator norm of the embedding satisfies
\begin{displaymath}
  \label{eq:EMB}
  \left\| [\cH]^\alpha \hookrightarrow L^{\infty}(\cX,\mu) \right\|_{\mathrm{op}} = M_{\alpha} < \infty.
\end{displaymath}
This embedding property was later summarized as an embedding index assumption (see, e.g., \citealt{li2023asymptotic}, Assumption 2) and led to the minimax optimality of spectral algorithms for $0<s<1$ \citep{zhang2024optimality}.
The embedding index assumption postulates that $\alpha_{0} = 1/\beta$, where $ \alpha_{0} $ is defined as
\begin{displaymath}
  \alpha_{0} = \inf\left\{ \alpha \in [1/\beta,1] :  \left\|[\cH]^\alpha \hookrightarrow L^{\infty}(\cX,\mu)\right\|_{\mathrm{op}} < \infty  \right\}.
\end{displaymath}

The minimax optimality-type results considered the upper bound of the generalization error of an algorithm and the algorithm-independent minimax lower bound.
In the renaissance of kernel regression arising from the study of neural networks, some recent work \citep{li2023asymptotic,Cui2021GeneralizationER,Bordelon2020SpectrumDL,li2024generalization} further considered the learning curve of kernel regression, which aims to obtain precise formulas or exact order of the generalization error (both upper and lower bounds) under any choice of the regularization parameter and even any noise level.
The learning curve-type results provide a nearly comprehensive picture of an estimator's generalization ability.
Note that in order to obtain a reasonable lower bound in the learning curve scenario, we need to assume that the source condition of $f^{*}$ is exactly $s$, i.e., $f^{*} \in [\cH]^{s},$ and $ f^{*} \notin [\cH]^{r}, \forall r > s$.
Specific descriptions of this condition were provided in these learning curve papers (see \citealt{Cui2021GeneralizationER}, Eq.(8); or \citealt{li2023asymptotic}, Assumption 3).

Next, we formally state the learning curve result for kernel gradient flow (from \citealt{li2024generalization}, Theorem 3.1).
Under Assumptions \ref{ass EDR} and \ref{ass source condition}, with $\beta >1, s>0$, and the embedding index assumption, by choosing $t \asymp n^{\theta}, \theta > 0$, we have
\begin{equation}\label{eq lc of kgf}
    \cR(\hat{f}_{t}) = \begin{cases} \Theta_{\PP}\left(n^{- s \theta} + n^{-(1-\theta / \beta)}\right), & \text { if } \theta<\beta, \\ \sigma_{0}^{2} \cdot \tilde{\Omega}_{\PP}\left(1\right), & \text { if } \theta \geq \beta,\end{cases}
\end{equation}
where $ a_{n} = \tilde{\Omega} (b_{n}) $ means $a_{n} = \Omega((\ln{n})^{-p} b_{n})$ for any $p>0$, and $ \tilde{\Omega}_{\PP} $ is the probability version of $ \tilde{\Omega}$.
In fact, \cite{li2024generalization} addressed all analytic spectral algorithms, including KRR and KGF.

\begin{figure}[htbp]
\centering
\begin{minipage}[b]{0.49\columnwidth}
\centering
\includegraphics[width=\linewidth]{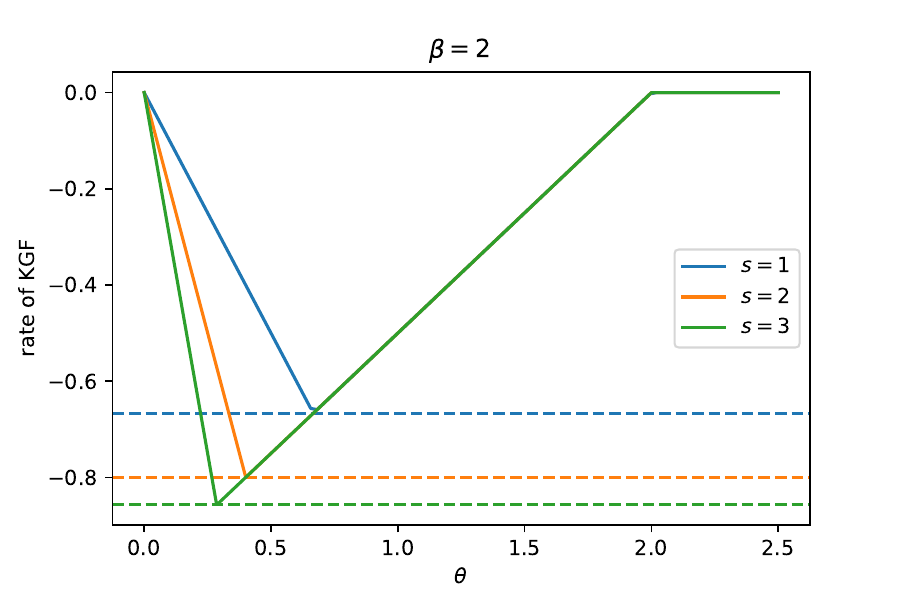}\\[-0.5ex]
\revtext{\textnormal{(a)}}
\end{minipage}\hfill
\begin{minipage}[b]{0.40\columnwidth}
\centering
\includegraphics[width=\linewidth]{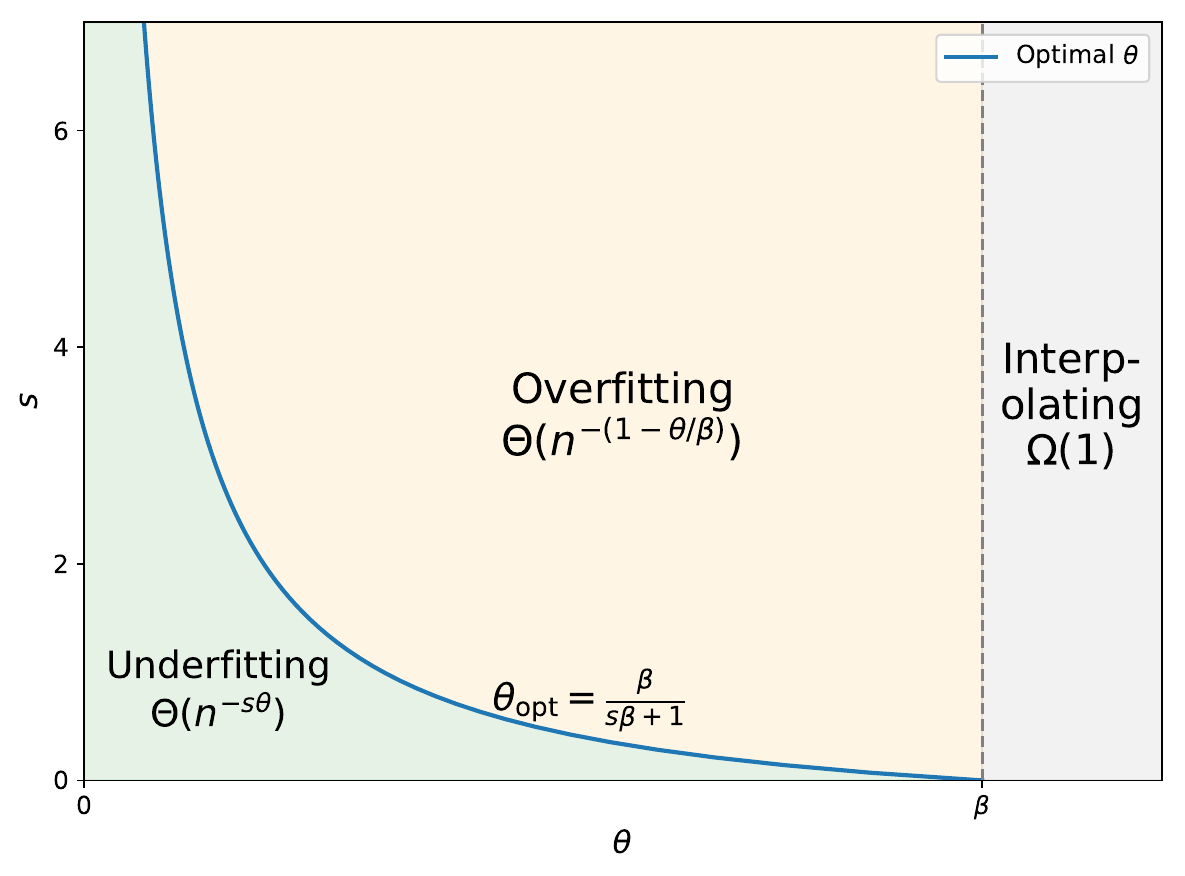}\\[-0.5ex]
\revtext{\textnormal{(b)}}
\end{minipage}
\caption{Asymptotic learning curve of kernel gradient flow.}
\label{figure learning curve}
\end{figure}

The first line of \eqref{eq lc of kgf} is the regularized regime, where the two terms correspond to the bias and variance terms, respectively.
As a direct corollary, choosing the optimal regularization $ t_{\text{opt}} \asymp n^{\frac{\beta}{s \beta + 1}}$ leads to the optimal convergence rate of the kernel gradient flow:
\begin{equation}\label{eq minimax rate}
    n^{-\frac{s \beta}{s \beta + 1}},
\end{equation}
which is also the minimax lower rate of the function space $[\cH]^{s}$.
The second line of \eqref{eq lc of kgf} is the interpolating regime, where the regularization is slight, and the estimator behaves similarly to the kernel interpolation estimator (see Section \ref{subsubsection KI fix d}).
The result implies that the generalization ability in the interpolating regime can be arbitrarily bad.
Figure \ref{figure learning curve} visualizes the convergence rates for different values of $\theta > 0: t \asymp n^{\theta},$ and source condition $s>0$.
The dashed lines in Figure \ref{figure learning curve} (a) represent the minimax rates achieved by the optimal regularization $t$.

\cite{Cui2021GeneralizationER} and \cite{li2023asymptotic} also discussed the learning curve of the kernel ridge regression and the impact of the noise level.
Compared with the convergence rates of KGF in \eqref{eq lc of kgf}, the difference in KRR is the saturation effect, while KGF never saturates.
The saturation effect, first conjectured by \cite{bauer2007_RegularizationAlgorithms}, says that when $s>2$, regardless of how carefully the KRR is tuned, the convergence rate cannot be faster than $ n^{-\frac{2 \beta}{2 \beta +1}}$, which is suboptimal compared with the minimax rate $ n^{-\frac{s \beta}{s \beta +1}}$.

To end this subsection, we consider an example of the Sobolev RKHS.
\revtext{Assume that $\cX \subset \RR^{d}$ is a bounded domain with a smooth boundary and that the input distribution $\mu$ on $\cX$ has Lebesgue density $p$ satisfying $0 < c \le p(x) \le C$ for two constants $c$ and $C$.}
The (fractional) Sobolev space $W^{m,2}(\cX)$ is an RKHS if $m > d/2$ \citep{adams2003_SobolevSpaces}.
Denoting $\cH = W^{m,2}(\cX)$, previous results have shown that the decay rate of the eigenvalues of $\cH$ is $\beta = 2m/d$ \citep{edmunds_triebel_1996}, and $\cH$ satisfies the embedding index assumption $\alpha_{0} = 1/\beta$.
Furthermore, the interpolation space of $\cH$ is still a Sobolev space, i.e., $ [\cH]^p \cong W^{mp,2}(\cX), \forall p>0 $.
Suppose that the true function $ f^{*} \in W^{r,2}(\cX)$ for some $r>0$ and that we use the kernel associated with $\cH = W^{m,2}(\cX) $ to run the kernel gradient flow.
Then, the source condition of $f^{*}$ with respect to $\cH$ is $s = r/m$.
Thus, we know that the optimal convergence rate of the generalization error is $ s \beta / (s \beta + 1) = 2r/(2r+d)$, which is consistent with the minimax rate of $ W^{r,2}(\cX) $.

\subsubsection{Kernel interpolation}\label{subsubsection KI fix d}

In practice, neural networks are usually trained to a near-zero training error and have good generalization ability \citep{Belkin2018ToUD,zhang2021understanding}.
This benign overfitting phenomenon of neural networks makes one wonder about the generalization ability of kernel interpolation estimator $\hat{f}_{\mathrm{inter}}$ in \eqref{eq KI estimator}, which is also the interpolation estimator with the minimum RKHS norm:
\begin{displaymath}
    \hat{f}_{\mathrm{inter}} = \mathop{\arg\min}\limits_{f \in \cH} \| f \|_{\cH}, ~\text{s.t.}~  f(\bx_i) = y_i, \quad i = 1,\ldots,n.
\end{displaymath}
In fixed dimensions, several works have claimed the inconsistency of $\hat{f}_{\mathrm{inter}}$ under various settings.
\cite{rakhlin2019consistency} showed that $\hat{f}_{\mathrm{inter}}$ with the Laplace kernel is inconsistent when dimension $d$ is odd.
\cite{buchholz2022kernel} proved the inconsistency when $k$ is the kernel function associated with the Sobolev space $ W^{m,2}(\cX), d/2 < m < 3d/4$.
\cite{beaglehole2023inconsistency} showed the inconsistency for a class of shift-invariant periodic kernels under mild spectral assumptions.
\cite{10.1093/biomet/asad048} showed that the generalization ability of kernel interpolation can be arbitrarily bad for those RKHSs that satisfy the embedding index assumption.
\cite{lai2023generalization} showed that kernel interpolation with the neural tangent kernel of a fully connected two-layer ReLU neural network on one-dimensional data is approximately a linear interpolation.
All of these results suggest that kernel interpolation can not generalize well in fixed dimensions.

Beyond inconsistency, \cite{mallinar2022benign} and \cite{cheng2024characterizing} discussed the tempered regime (the generalization error remains bounded) and the catastrophic regime (the generalization error diverges to infinity) of kernel interpolation.
\cite{haas2024mind} found that adding spike components to kernels could lead to consistent or even rate-optimal kernel interpolation in fixed dimensions.

\subsection{Kernel regression in high dimensions}\label{subsection large d kr}

Since neural networks often perform well on high-dimensional data, high-dimensional kernel regression has garnered much recent interest.
In this subsection, we still use the notations in Section \ref{subsection preli of kr} and summarize the results when the sample size and dimension satisfy $n \asymp d^{\gamma}$ for some $\gamma > 0$.
Compared with the fixed-dimensional setting, as $d$ varies, the eigenvalues of $\cH$ usually depend on $d$ in an unpleasant way.
Thus, the polynomial decay rate of the eigenvalues in Assumption \ref{ass EDR} must not hold.
Most existing results considered specific kernels (e.g., inner product kernel) or special input spaces (e.g., sphere, discrete hypercube), where the eigenvalues and eigenfunctions of $\cH$ are well understood.
Nevertheless, many new phenomena have emerged in high dimensions.

\subsubsection{Polynomial approximation barrier}\label{subsubsection poly barrier}

We first review the ``polynomial approximation barrier'' of kernel regression studied by \cite{Ghorbani2019LinearizedTN} and several subsequent publications \citep[etc.]{Ghorbani_When_2021,mei2022generalization,Ghosh_three_2021,xiao2022precise,hu2022sharp,misiakiewicz_spectrum_2022,Donhauser_how_2021,mei2021learning}.
This line of work assumed the true function $ f^{*}$ to be square-integrable.
Specifically, \cite{Ghorbani2019LinearizedTN} considered the inner product kernel on the sphere $\cX = \mathbb{S}^{d-1}$ with uniform distribution, defined as
\begin{equation}\label{eq inner product kernel}
    k(\bx, \bx^{\prime}) = h\left( \langle \bx, \bx^{\prime} \rangle\right), ~\forall \bx, \bx^{\prime} \in \mathbb{S}^{d-1},
\end{equation}
where $ h(z) \in \mathcal{C}^{\infty} \left([-1,1]\right)$ is a fixed function independent of $d$ and
    \begin{displaymath}
        h(z) = \sum_{j=0}^\infty a_j z^j, ~ a_{j} > 0, ~\forall j = 0, 1, 2,\ldots.
    \end{displaymath}
(This definition of kernel is equivalent to assuming that Assumption 3 of \cite{Ghorbani2019LinearizedTN} holds for all levels $\ell \in \NN$.)
The eigenfunctions of such kernels are spherical harmonic polynomials, and there exists a concise characterization of the order of eigenvalues (see, e.g., \citealt{smola2000_RegularizationDotproduct} and \cite{lu2023optimal}).
When $n \asymp d^{\gamma}, \gamma \in (0,\infty) \backslash \NN^{+}$, Theorem 4 of \cite{Ghorbani2019LinearizedTN} showed that the generalization error of the kernel ridge regression $\hat{f}_{\lambda}$ with $\lambda=O(n^{-1})$ satisfies (with high probability)
\begin{displaymath}
    \left| \cR(\hat{f}_{\lambda}) -\left\|\mathbf{P}_{>\ell} f^{*}\right\|_{L^2}^2\right| \leq \varepsilon\left(\left\|f^{*}\right\|_{L^2}^2+\sigma_{0}^{2}\right),
\end{displaymath}
where $\ell = \lfloor \gamma \rfloor$, $\mathbf{P}_{\le \ell}$ denotes the projection operator that projects to the subspace of polynomials of degree at most $\ell$,  $\mathbf{P}_{>\ell} = \mathbf{I} -\mathbf{P}_{\le \ell}, $ and $\varepsilon$ is a positive real number.
The results can be viewed more intuitively as
 \begin{displaymath}
     \cR(\hat{f}_{\lambda}) =  \| \mathbf{P}_{>\ell} f^{*}\|_{L^2}^2 + o_{\PP}(1).
 \end{displaymath}
They also showed that $ \| \mathbf{P}_{>\ell} f^{*}\|_{L^2}^2 + o_{\PP}(1)$ is the best generalization error achievable by a kernel regression in the form of $\hat{f}(\bx) = \sum_{i=1}^{n} a_{i} k(\bx,\boldsymbol{x_{i}}), a_{i} \in \RR$.
This polynomial approximation barrier was later used to quantify advantages of feature learning in some literature (see Section \ref{attemps for feature learning}).
In the case of $\gamma \in \NN^{+}$, which was not covered by \cite{Ghorbani2019LinearizedTN}, subsequent works \cite{xiao2022precise}, \cite{hu2022sharp}, and \cite{misiakiewicz_spectrum_2022} derived the precise asymptotic formulas for the bias and variance terms of the kernel ridge regression and showed that the generalization error achieved the peak when $ n \approx d^{\gamma}/ \gamma!, \gamma \in \NN^{+}$.
\cite{mei2022generalization} studied a similar approximation barrier for kernel regression for kernels whose eigenspaces have hypercontractivity and satisfy certain spectral conditions.

The key idea of this line of work is to decompose the empirical kernel matrix $ \KK(\bX,\bX)$ into low-frequency and high-frequency parts, then to take advantage of the nice properties of the eigenfunctions (which are spherical harmonic polynomials for the inner kernel of the product on the sphere) when $d$ is large enough.

\subsubsection{Generalization behaviors under the source condition}\label{subsection optimality under source}
Another line of work considered the source condition (Assumption \ref{ass source condition}) in the high-dimensional setting \citep{Liu_kernel_2021,zhang2024optimal,lu2023optimal}.
Compared with the work mentioned in Section \ref{subsubsection poly barrier}, which only assumes $f^{*}$ to be square-integrable, $[\cH]^{s}$ is a smaller function space than $L^{2}(\cX,\mu)$ when $s>0$, and $L^{2}(\cX,\mu)$ can be viewed as a limiting case of $s \to 0$.

\begin{figure}[htbp]
\centering
\includegraphics[width=0.45\columnwidth]{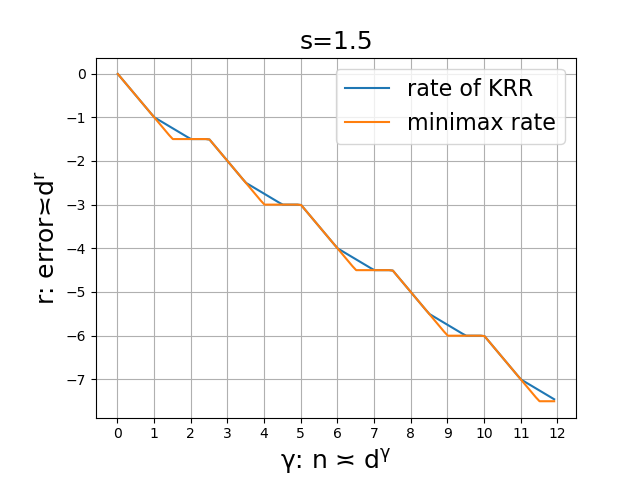}
\caption{Best convergence rates of KRR and corresponding minimax lower rate (w.r.t. $d$) for $s=1.5$ and $\gamma>0$.}
\label{figure rate of d}
\end{figure}

Next, we review the results in \cite{zhang2024optimal}, which also studied the inner product kernel on the sphere \eqref{eq inner product kernel}, provided minimax optimal rates, and found some new phenomena for the kernel ridge regression.
Note that two parameters, $s$ and $\gamma$, determine the generalization ability.
\cite{zhang2024optimal} showed that when $s \in (0,1)$, $s \in [1,2]$ and $s>2$, the generalization error behaved differently along $\gamma$ (\citealt{zhang2024optimal}, Theorems 2 and 3).
Specifically, when $0<s<1$, the generalization error under the best choice of the regularization parameter (denoted as $R^{*}$) has two periods:
\begin{itemize}
\item [(i)] if $\gamma \in \left(p+ps,p+ps+s \right]$, $p \in \NN$,
    \begin{displaymath}
         R^{*}=\Theta_{\PP}\left( d^{-\gamma + p} \right) = \Theta_{\PP}\left( n^{-1 + \frac{p}{\gamma}} \right);
    \end{displaymath}

\item [(ii)] if $\gamma \in \left(p+ps+s,(p+1)+(p+1)s \right]$, $p \in \NN$,
    \begin{displaymath}
        R^{*} = \Theta_{\PP}\left( d^{-(p+1)s} \right) = \Theta_{\PP}\left( n^{- \frac{(p+1)s}{\gamma}} \right).
    \end{displaymath}
\end{itemize}
When $1 \le s \le 2$, $R^{*}$ has three periods:
\begin{itemize}
\item [(i)] if $\gamma \in \left(p+ps,~ p+ps+1 \right]$, $p \in \NN$,
    \begin{displaymath}
         R^{*}=\Theta_{\PP}\left( d^{-\gamma + p} \right) = \Theta_{\PP}\left( n^{-1 + \frac{p}{\gamma}} \right);
    \end{displaymath}

\item [(ii)] if $\gamma \in \left(p+ps+1,~ p+ps+2s-1 \right]$, $p \in \NN$,
    \begin{displaymath}
        R^{*} = \Theta_{\PP}\left( d^{-\frac{\gamma-p+ps+1}{2}} \right) = \Theta_{\PP}\left( n^{- \frac{\gamma-p+ps+1}{2 \gamma}} \right).
    \end{displaymath}

\item [\revtext{(iii)}] if $\gamma \in \left(p+ps+2s-1, (p+1)+(p+1)s \right]$, $p \in \NN$,
    \begin{displaymath}
        R^{*} = \Theta_{\PP}\left( d^{-(p+1)s} \right) = \Theta_{\PP}\left( n^{- \frac{(p+1)s}{\gamma}} \right).
    \end{displaymath}
\end{itemize}
When $s>2$, $R^{*} $ is exactly the same as $s=2$.
Furthermore, \cite{zhang2024optimal}, Theorem 5, provided the corresponding minimax lower rate of $[\cH]^{s}, s>0$, which is omitted here.
For instance,  Figure~\ref{figure rate of d} shows the best convergence rates of KRR and the corresponding minimax lower rates (with respect to $d$) for $s=1.5$ and $\gamma>0$.
More visualizations of these convergence rates can be found in \cite{zhang2024optimal}, Figures 2 and 3.

Several interesting phenomena arise from these results.
\begin{itemize}
\item Periodic plateau behavior: when $\gamma$ varies within a certain period, the rate with respect to dimension $d$ does not change with $\gamma$.
\item Multiple descent behavior: the rates with respect to sample size $n$ achieve peaks and isolated valleys at certain values of $\gamma$ (\citealt{zhang2024optimal}, Figure 3).
\item Minimax optimality and new saturation effect: when $0<s\le1$, the best convergence rate of KRR matches the minimax lower rate for all $\gamma > 0$.
When $s>1$, KRR cannot achieve the minimax lower rate for certain ranges of $\gamma$, which is called the new saturation effect of KRR.
\end{itemize}
The periodic plateau and multiple-descent behaviors were first reported in \cite{lu2023optimal} for kernel gradient flow with $s=1$.
They also included an example of the neural tangent kernel of the two-layer fully connected ReLU neural network.

\subsubsection{Kernel interpolation}\label{subsubsection KI large d}
We have shown in Section \ref{subsubsection KI fix d} that kernel interpolation $\hat{f}_{\mathrm{inter}}$ cannot generalize in fixed dimensions.
However, kernel interpolation can generalize surprisingly well in high dimensions, as first demonstrated theoretically by \cite{Liang_Just_2019}.
\cite{Liang_Just_2019} studied the inner product kernel and the setting $n \asymp d$.
Using the linear approximation of high-dimensional kernel matrices from \cite{Karoui_spectrum_2010}, \cite{Liang_Just_2019} first proved the upper bounds of the bias and variance terms of kernel interpolation, and showed that the generalization error will converge to zero when the data exhibit a low-dimensional structure.

\cite{liang2020multiple} also considered the inner product kernel and assumed that the coordinates of the $d$-dimensional input are independent and identically distributed.
Considering the high-dimensional setting $n\asymp d^{\gamma}, \gamma >0$, they proved an upper bound of the variance term.
Furthermore, assuming $ f^{*}(\bx) = \langle k(\bx,\cdot), \rho(\cdot)\rangle_{L^{2}}$, with $ \| \rho \|_{L^{4}}^{4} \le C$ for some constant $C>0$, they demonstrated that the bias term is infinitesimal of a higher order compared with the variance term.
They obtained the following upper bound for the generalization error with a concrete convergence rate (denote $l = \lfloor \gamma \rfloor$):
\begin{displaymath}
    \cR(\hat{f}_{\mathrm{inter}}) = O_{\PP} \left( d^{l-\gamma} + d^{\gamma-l-1} \right).
\end{displaymath}
This upper bound exhibits multiple-descent behavior, i.e., the convergence rate is non-monotone as $\gamma$ increases.
They also provided visualization and empirical evidence for the multiple-descent behavior (see \citealt{liang2020multiple}, Figures 1 and 2).

To our knowledge, \cite{aerni2022strong} is the first to provide lower bounds of the bias and variance terms for kernel interpolation in high dimensions.
They considered the convolutional kernel on the discrete hypercube $\{-1, 1\}^{d}$ and a special form of the true function $f^{*}(x) = x_{1}x_{2}\cdots x_{L^{*}}$, where $ L^{*}$ is formulated in \cite{aerni2022strong}, Theorem 1.
Using a similar decomposition of the empirical kernel matrix $\KK(\bX,\bX)$ (as mentioned at the end of Section \ref{subsubsection poly barrier}) as the line of work \cite{Ghorbani2019LinearizedTN}, this $f^{*}$ will fall into the eigenspace corresponding to the low-frequency part.
They demonstrated that kernel interpolation has generalization ability in this setting and discovered the multiple-descent behavior of kernel interpolation.

\begin{figure}[htbp]
\centering
\includegraphics[width=0.45\columnwidth]{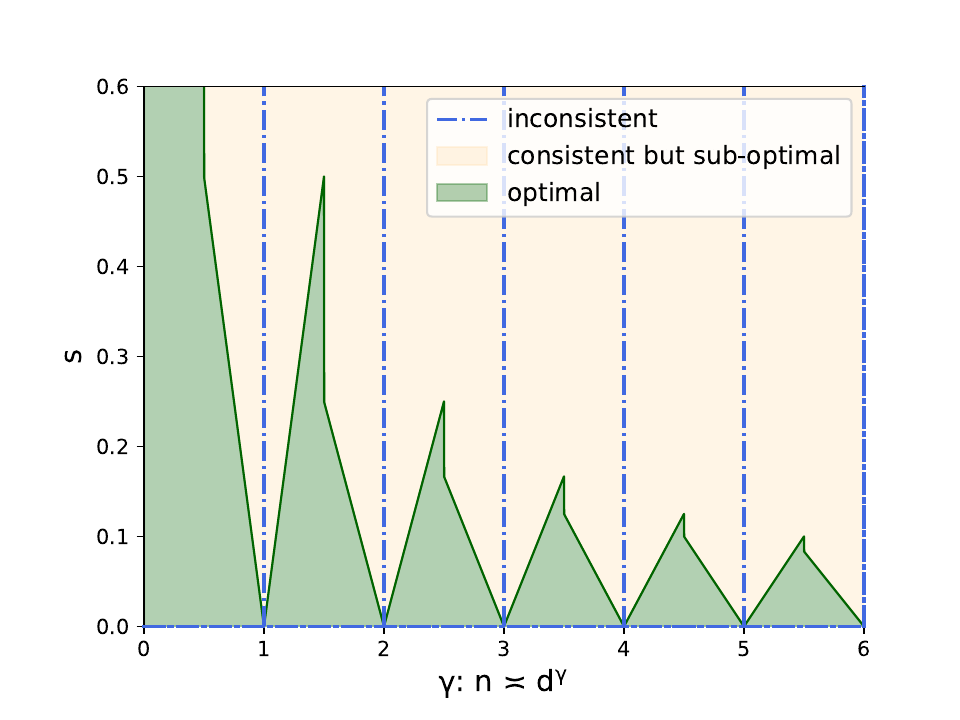}
\caption{Phase diagram about the consistency and the optimality of kernel interpolation.}
\label{figure phase diagrame}
\end{figure}

Furthermore, \cite{zhang2024phase} showed that whether kernel interpolation is a good choice (in the sense of minimax optimality) depends on the relative smoothness of the true function.
Specifically, they considered the inner product kernel on the sphere, $n \asymp d^{\gamma}, \gamma >0$, and assumed the source condition of $f^{*}$ to be exactly $s \ge 0$.
For all values of $s \ge 0$ and $\gamma > 0$, they fully characterized the exact orders of the bias and variance terms, leading to an exact order of the generalization error: (denote $l = \lfloor \gamma \rfloor$)
\begin{displaymath}
    \cR(\hat{f}_{\mathrm{inter}}) = \Theta_{\PP} \left( d^{l-\gamma} + d^{\gamma-l-1} + d^{-(l+1)s} \right).
\end{displaymath}
Comparing this rate with the minimax lower rate in $ [\cH]^{s} $ in \cite{zhang2024optimal}, they further showed that for different values of $(s,\gamma)$, kernel interpolation can be minimax optimal, consistent but sub-optimal, or inconsistent (see Figure~\ref{figure phase diagrame}, borrowed from \citealt{zhang2024phase}).
Specifically, for any fixed $\gamma \in (0,\infty) \backslash \NN^{+}$, there exists a threshold $\Gamma(\gamma)$ (see \citealt{zhang2024phase}, Eq.(11)) such that when $s > \Gamma(\gamma)$, kernel interpolation is sub-optimal; when $ 0 < s \le \Gamma(\gamma)$, kernel interpolation is minimax optimal; and when $s = 0$ or $\gamma \in \NN^{+}$, kernel interpolation is inconsistent.
The existence of a threshold $\Gamma(\gamma)$ provides a comprehensive answer to the question ``when does the benign overfitting phenomenon occur'' in kernel regression and shows how it depends on the relative smoothness of the true function and the high-dimensional scaling.

\subsubsection{Other topics}

Recall that most of the results discussed above focused on specific kernels and input spaces.
\begin{revision}
\citet{barzilai2023generalization} and \citet{misiakiewicz2024non} aimed to provide a unified theory of the generalization error of kernel regression under mild assumptions.
In addition to general kernel ridge regression bounds, \citet{gavrilopoulos2024geometrical} analyze a conjugate kernel obtained after training its feature map on the data; we compare this mechanism with the adaptive feature model in Section \ref{subsection nn as ada kernel}.
\end{revision}
It is also worth mentioning the study of benign overfitting in high-dimensional linear regression with an arbitrary structure of the covariance of the input data \citep[etc.]{bartlett2020benign,tsigler2023benign,10.1214/21-AOS2133,kobak2020optimal}.
It can be seen that high-dimensional linear regression and kernel regression share many common characteristics.
 \section{\revheading{From fixed features to feature learning}}\label{section from fix to feature}

\subsection{\revheading{Limitations of fixed features}}\label{subsection limitations ntk}

Despite significant successes of the neural tangent kernel (NTK) theory, an increasing number of researchers argue that it does not explain the outstanding performance of real-world neural networks.
Notable works highlighting these limitations include, but are not limited to, \cite{wei2019regularization} and \cite{ghorbani2019limitations,Ghorbani_When_2021} from the theoretical perspective; \cite{chizat2019lazy} and \cite{arora2019exact} from the empirical perspective.
The most widely discussed shortcoming of the NTK theory is its lack of feature learning capability, since it equates the training dynamics of neural networks entirely with kernel regression using the specific neural tangent kernel.
In the following, we analyze the limitations of the NTK theory from multiple perspectives.

For a kernel function $k$, Mercer's decomposition \eqref{eq mercer decom} can be written as
\begin{revision}
\begin{displaymath}
    k(\bx,\bx^{\prime}) = \Psi(\bx)^{\top} \bLambda \Psi(\bx^{\prime}),
\end{displaymath}
\end{revision}
where we denote $\Psi(\bx) = (\psi_{1}(\bx), \psi_{2}(\bx), \ldots)^{\top}$ and $\bLambda = \text{Diag}\{ \lambda_{1}, \lambda_{2},\ldots \}$.
Define the feature map:
\begin{equation}\label{def of feature map}
    \bx \to \Phi(\bx) = \bLambda^{\frac{1}{2}} \Psi(\bx).
\end{equation}
\revtext{The kernel function is then the inner product of the two features at $\bx$ and $\bx^{\prime}$, i.e.,}
\begin{equation}\label{eq k equals inner of feature}
k(\bx,\bx^{\prime}) = \Phi(\bx)^{\top} \Phi(\bx^{\prime}).
\end{equation}
It is well known that kernel regression is equivalent to a linear regression in the feature space.
Specifically, (i) the kernel ridge regression estimator in Definition \ref{def of krr} is the same as the ridge regression estimator in the feature space; (ii) the kernel gradient flow estimator in Definition \ref{def of kgf} can be expressed as $ \hat{f}_{t}(\bx) = \Phi(\bx)^{\top} \boldsymbol{b}_{t} $, where $\boldsymbol{b}_{t}$ is obtained by a gradient flow to minimize square loss $\cL(\boldsymbol{b})=\|\bY - \Phi(\bX)^{\top} \boldsymbol{b}\|_{2}^{2}/n$.
Once the architecture of a neural network is determined, the feature corresponding to the NTK is fixed and independent of the data.
Therefore, kernel regression with NTK disregards neural networks' ability to learn features from the data.

\begin{revision}
In the classical nonparametric regimes reviewed in Section \ref{subsection lc fix d}, an appropriately regularized fixed kernel method can attain the minimax rate when the kernel and source condition are matched.
However, the minimax rate is a worst-case benchmark over a function class and need not coincide with the best achievable rate for a specific $f^{*}$.
When a pre-specified feature map is poorly aligned with $f^{*}$, as in the high-dimensional polynomial approximation barrier of Section \ref{subsubsection poly barrier}, feature learning can offer an advantage if training changes the representation to improve that alignment.
\end{revision}

Additionally, the results in Sections \ref{subsubsection KI fix d} and \ref{subsubsection KI large d} show that kernel interpolation cannot generalize in fixed dimensions and does not outperform regularized kernel regression (unless $f^{*}$ is less smooth) in high dimensions.
This also suggests that kernel regression may not be an appropriate method to interpret ``the benign overfitting phenomenon '', which is widely observed for neural networks \citep{zhang2021understanding}.

In the NTK theory, a special symmetric initialization of the parameters is widely used \citep{chizat2019lazy,hu2019simple,lai2023generalization}, ensuring that the output of a neural network at initialization is zero, e.g., $ f_{\btheta(0)}^{m}(\bx) \equiv 0$ in \eqref{eq relu two bias}.
Recall that the explicit expression of the KGF estimator \eqref{eq explicit expression} also relies on the zero initialization $ \hat{f}_{0}(\bx) \equiv 0$.
Recent work \citep{Chen2024OnTI} showed that, in the NTK regime, nonzero initialization of a neural network would introduce a terrible bias.
Specifically, they showed that a sufficiently wide and fully connected neural network with each parameter initialized as an independent standard normal variable achieves an optimal convergence rate of generalization error $ n^{-\frac{3}{d+3}}$, no matter how smooth $f^{*}$ is (in the sense of the source condition $f^{*} \in [\cH]^{s}$).
This convergence rate falls into the ``curse of dimensionality'': As the dimensionality $d$ increases, the sample size $n$ required for good generalization ability increases exponentially.
Thus, they suggested that there is a divergence between the NTK theory and real-world neural networks.

\subsection{\revheading{Implications of fixed features}}\label{subsection implications of fixed kernel}
\revtext{Although the theory of fixed features has limitations in explaining real-world neural networks, it offers insights into what constitutes a good feature representation.}
In this subsection, as a warm-up of the adaptive feature approach, we review two examples to illustrate a key observation:
\begin{quote}
\textit{The alignment between the true function and the chosen feature matters.}
\end{quote}
\begin{revision}
More precisely, for a given kernel feature map $\Phi(\bx)$ (defined in \eqref{def of feature map}) and the corresponding kernel $k$, we say that the feature map is more aligned with the true function $f^{*}$ when a larger share of the squared projection energy $\langle f^{*}, \psi_{j}\rangle_{L^{2}}^{2}$ lies in eigenspaces associated with larger eigenvalues $\lambda_{j}$.
This is a property of a feature representation fixed before training.  For the high-dimensional inference problems of \citet{arous2021online}, the information exponent governs the sample complexity of an initial phase in which online SGD obtains nontrivial correlation with the target.  \citet{abbe2024merged,abbe2023leap} study related two-layer SGD settings in which learnability depends on hierarchical structure in the target.  By contrast, \citet{lecue2025sharp} formalize alignment and feature learning through a feature space decomposition: the spectral methods they study can exploit alignment in a fixed representation but do not learn a feature map that depends on the target.
Thus, alignment helps a fixed kernel only when the relevant directions are already available in its feature space, whereas feature learning concerns how training acquires useful correlation with the target.
\end{revision}

\subsubsection{\revheading{An alignment example with two kernels}}
Consider the following example of a one-dimensional regression problem: suppose that $\cX=[0,1]$ with uniform distribution.
Consider a true function $f^{*}(x) = \sin(2 \pi x)$, and apply kernel regression with the following two kernels:
\begin{displaymath}
    k_{1}(x, x^{\prime})=\mathbf{1}_{\{x \geq x^{\prime}\}}(1-x) x^{\prime}+\mathbf{1}_{\{x<x^{\prime}\}}(1-x^{\prime}) x,
\end{displaymath}
and
\begin{displaymath}
    k_{2}(x, x^{\prime})=\min\{x,x^{\prime}\}.
\end{displaymath}
The results of \cite{wainwright2019_HighdimensionalStatistics} and \cite{li2022saturation} give rise to explicit formulas for the eigenvalues and eigenfunctions of $k_{1}, k_{2}$, i.e., for $j=1,2,\ldots$
\begin{displaymath}
    \lambda_{1,j}=\frac{1}{\pi^2 j^2}, \quad \psi_{1,j} =\sqrt{2} \sin (j \pi x),
\end{displaymath}
and
\begin{displaymath}
    \lambda_{2,j}= \frac{4}{\pi^{2}(2j-1)^{2}}, ~~ \psi_{2,j}(x)=\sqrt{2} \sin \left(\frac{2 j-1}{2} \pi x\right).
\end{displaymath}
Therefore, $k_{1}, k_{2}$ have the same decay rates of eigenvalues, $ \beta_{1} = \beta_{2} = 2$.
Recalling the definition of an interpolation space $[\cH]^{s}$ in \eqref{def of Hs}, since the projection $\langle f^{*}, \psi_{1,j}\rangle $ is nonzero only when $ j =2$, $f^{*}$ has the source condition $ s_{1} = \infty $ with respect to $k_{1}$.
Since the projections $\langle f^{*}, \psi_{2,j}\rangle \asymp j^{-2} $, $f^{*}$ has the source condition $ s_{2} = 1.5 $ with respect to $k_{2}$.
Using the results \eqref{eq minimax rate} in Section \ref{subsection lc fix d}, we obtain the optimal convergence rates as $r_{1} = n^{-1}, r_{2} = n^{-0.75}$, respectively, when using $k_{1}, k_{2}$ to estimate $f^{*}$.
This implies that the feature $\Phi_{1}$ (or kernel $k_{1}$), which aligns more with $f^{*}$, is a better choice than $\Phi_{2}$ (or $k_{2}$).

\subsubsection{\revheading{Alignment under the neural tangent kernel}}
The second example, from \cite{arora2019fine}, considered the two-layer fully connected ReLU neural network.
\revtext{Denote the noiseless labels as $\bY = (y_{1}, \ldots, y_{n})^{\top}$, and let the empirical NTK matrix $\KK \in \RR^{n \times n}$ have entries $\KK_{ij} = k(\bx_{i},\bx_{j})$, $i,j=1,\ldots,n$, where $k$ is the NTK of the fully connected two-layer ReLU neural network.}
Then under certain assumptions (the width of the hidden layer grows sufficiently fast, and the neural network is trained for a sufficiently long time), the generalization error of the neural network estimator $\hat{f}_{\mathrm{NN}}$ has an upper bound:
\begin{equation}\label{eq arora upper bound}
    \cR(\hat{f}_{\mathrm{NN}}) = O_{\PP}\left( \sqrt{\frac{\bY^{\top} \KK^{-1} \bY}{n}} \right).
\end{equation}
Let the eigenvalues and eigenvectors of the matrix $\KK$ be $ \hat{\lambda}_{j}, \hat{\boldsymbol{v}}_{j}$, i.e., $ \KK = \sum_{j=1}^{n} \hat{\lambda}_{j} \hat{\boldsymbol{v}}_{j} \hat{\boldsymbol{v}}_{j}^{\top}$, we have
\begin{equation}\label{eq arora decom}
    \bY^{\top} \KK^{-1} \bY = \sum\limits_{j=1}^{n} \frac{1}{\hat{\lambda}_{j}} \left( \hat{\boldsymbol{v}}_{j}^{\top} \bY\right)^{2}.
\end{equation}
$ \hat{\lambda}_{j}, \hat{\boldsymbol{v}}_{j} $ are empirical versions of the eigenvalues and eigenfunctions of $k$.
Also note that they considered the noiseless case, so $\bY$ is an empirical version of the true function $f^{*}$.
Therefore, \eqref{eq arora upper bound} and \eqref{eq arora decom} imply that projections $\langle f^{*}, \psi_{j} \rangle_{L^{2}}$ corresponding to small eigenvalues $\lambda_{j}$ must be small for a good generalization ability.

\subsection{Recent advances in feature learning}\label{attemps for feature learning}

The aim of the adaptive feature approach in the next section is to theoretically analyze the feature learning process of neural networks and its advantages.
Before that, we summarize some of the recent advances in feature learning.

There is a line of work empirically studying the evolution of the alignment between labels and features during the training of neural networks \citep[etc.]{oymak2019generalization,kopitkov2020neural,fort2020deep,maennel2020neural,shan2021theory,baratin2021implicit,atanasov2021neural,ortiz2021can}.
Some of these papers considered an index called ``(centered) kernel alignment'' \citep{cortes2012algorithms,kornblith2019similarity} under various settings and observed that the alignment increases as training progresses.
\begin{revision}
Moreover, even for neural networks with infinite width,
the parameterization, initialization scale, and learning rate scaling jointly determine whether the training dynamics remain in a kernel regime or exhibit feature learning.
In particular, the mean field limits \citep{chizat2019lazy,geiger2020disentangling,woodworth2020kernel,yang2020feature,bordelon2022self} show that the alignment can still increase during training.
\end{revision}
In these scenarios, feature learning refers to the phenomenon that the parameters of neural networks evolve non-trivially and cannot be considered approximately unchanged, which is in contrast to the frozen feature in the neural tangent kernel (NTK) regime.

Theoretically, characterizing what feature can be learned from the data and analyzing its impact on the method's generalization ability is a challenge.
\revtext{One line of work studies recursive feature machines to provide an explicit adaptive kernel construction: in the fully connected and convolutional settings studied by \citet{radhakrishnan2022mechanism,beaglehole2023mechanism}, the feature map is updated from the data rather than held fixed as in an NTK approximation.}
A notable line of work dealt with the problem by considering the one-step gradient descent setting \citep{ba2022high,damian2022neural,dandi2023learning,moniri2023theory,cui2024asymptotics}.
These works originated from the random feature model \citep{rahimi2007random, gerace2020generalisation, mei2022generalization2, hu2022universality}, etc., and aimed to prove that feature learning brought by one-step gradient descent can lead to advantages over random features and kernel regression estimators.
Next, we briefly introduce the results in \cite{ba2022high}.
They considered the following two-layer fully connected neural network
\begin{equation}\label{eq relu wo bias}
    f_{\mathrm{NN}}(\bx)=\frac{1}{\sqrt{m}} \sum_{r=1}^m a_r \sigma\left( \boldsymbol{w}_{r}^{\top} \bx \right)=\frac{1}{\sqrt{m}} \boldsymbol{a}^{\top} \sigma\left(\bW^{\top} \bx\right),
\end{equation}
and the proportional asymptotic limits, i.e., the width $m$, sample size $n$ and sample dimension $ d $ all tend to infinity and satisfy
\begin{displaymath}
n / d \rightarrow \gamma_1, m / d \rightarrow \gamma_2, ~~ \gamma_1, \gamma_2 \in(0, \infty).
\end{displaymath}
Denoting $ \bW_{0} $ as the weight matrix at initialization, they first updated $ \bW_{0} $ for one step:
\begin{displaymath}
    \bW_{1} = \bW_{0} + \eta \sqrt{m} \cdot \boldsymbol{G}_{0},
\end{displaymath}
where $ \boldsymbol{G}_{0} $ is the gradient of the square loss with respect to $ \bW_{0}$, and $\eta$ is the learning rate.
Then, a ridge regression was conducted in the new feature space $ \bx \to \sigma\left( \bW_{1}^{\top} \bx \right)$.
They also assumed that the true function was a single index function $f^{*}(\bx) = \sigma^{*}(\langle \bx, \boldsymbol{\beta_{*}}  \rangle)$, where $\bx \sim \mathcal{N}(\boldsymbol{0},\mathbf{I}_{d})$ and $ \| \boldsymbol{\beta_{*}} \|_{2} = 1 $.
Denote $\cR_{0}(\lambda) $ and $ \cR_{1}(\lambda)$ as the generalization errors of the ridge regression estimators with the regularization parameter $\lambda > 0$ using the features $ \sigma\left( \bW_{0}^{\top} \bx \right) $ and $  \sigma\left( \bW_{1}^{\top} \bx \right)$, respectively.
Under some detailed assumptions, by choosing the learning rate $\eta = \Theta(1)$, they proved that (\citealt{ba2022high}, Theorem 5):
\begin{displaymath}
    \cR_{0}(\lambda) - \cR_{1}(\lambda) \xrightarrow{\PP} \delta \geq 0,
\end{displaymath}
where $ \delta $ is constant with explicit expression in their paper.
For this constant learning rate $\eta = \Theta(1)$, one-step gradient descent has already shown an improvement over the initial random feature.
Furthermore, by choosing a larger learning rate $\eta = \Theta(\sqrt{m})$, they proved an upper bound of $ \cR_{1}(\lambda)$ (\citealt{ba2022high}, Theorem 7), which outperforms the kernel lower bound $ \left\|\mathbf{P}_{>1} f^*\right\|_{L^2}^2  $ (\citealt{hu2022universality,montanari2022interpolation}, the same as the polynomial approximation barrier in Section \ref{subsubsection poly barrier}) for some examples.

Something more interesting was also studied about the first step gradient $\boldsymbol{G}_{0}$ and the new feature $\bW_{1}$.
They showed that $\boldsymbol{G}_{0}$ is close to a rank-1 matrix $ \bX^{\top} \bY \boldsymbol{a}^{\top}/n\sqrt{m} $ (omitting the constant), where $\boldsymbol{a}$ is the weights of the randomly initialized output layer, and $\bX^{\top} \bY$ is roughly $\bbeta_{*}$ in their true function.
Furthermore, they showed that the first singular vector $\boldsymbol{u}_{1}$ of $\bW_{1}$ (corresponding to the leading singular value) satisfies the following:
\begin{displaymath}
    \left|\left\langle\boldsymbol{u}_1, \bbeta_*\right\rangle \right|^2 \xrightarrow{\PP} C_{0},
\end{displaymath}
where the expression $C_{0} \in (0,1)$ is provided in their paper.
It can be seen from their expression that $C_{0} \to 1$ when $\gamma_{1} \to \infty$, and $C_{0}$ increases as the learning rate increases.
Despite many specific assumptions, they theoretically proved that the alignment (similar to the alignment discussed at the beginning of Section \ref{subsection implications of fixed kernel}) between the feature and labels emerges from the training of neural networks, and showed the advantage of feature learning.

There is a substantial body of literature studying feature learning, which is too extensive to list comprehensively here.
For instance, \cite{hanin2019finite}, \cite{dyer2019asymptotics}, \cite{huang2020dynamics}, \cite{yaida2020non}, \cite{naveh2021self}, and \cite{bordelon2024dynamics} studied the finite-width corrections which enhanced feature evolving.
Toward analyzing feature learning with fewer assumptions on the true function, \cite{radhakrishnan2022mechanism}, \cite{beaglehole2023mechanism}, \cite{beaglehole2024gradient}, and \cite{radhakrishnan2024mechanism} proposed a general structure of the weights of neural networks during training, which was called the neural feature ansatz.
 \begin{revision}
\section{Adaptive feature model and over-parameterized Gaussian sequence}\label{section over-para gsm}

\subsection{Neural networks as adaptive feature model}\label{subsection nn as ada kernel}

As discussed in Sections \ref{subsection limitations ntk} and \ref{subsection implications of fixed kernel}, kernel regression is equivalent to linear regression in its associated feature space, and the alignment between the feature and the target function affects generalization.
Neural networks can likewise be viewed as linear regressions on features that are learned during training.
Throughout this section, we consider one-dimensional outputs.
Let $\bbeta \in \RR^{m}$ denote the last-layer weights and $\Phi_{\bEta}(\bx) \in \RR^{m}$ the feature before the last layer.
The network output then has the form
\begin{equation}\label{eq nn as linear in feature}
  f_{\btheta}(\bx) = \Phi_{\bEta}(\bx)^{\top} \bbeta,
\end{equation}
where $\btheta = \text{vec}(\{ \bbeta,\bEta\})$ collects the learnable parameters of the neural network.
For example, in \eqref{eq relu wo bias}, $\bbeta = \boldsymbol{a}$, $\Phi_{\bEta}(\bx) = \sigma(\bW^{\top}\bx)/\sqrt{m}$, and $\bEta = \bW$.

Although the parameterization of $\Phi_{\bEta}(\bx)$ is architecture dependent and can be complicated, the estimator in \eqref{eq nn as linear in feature} admits an abstract representation as a more general \textit{adaptive feature model}.

\begin{definition}[Adaptive feature model]\label{def ada kernel model}
  Consider the nonparametric regression problem \eqref{main data model}.
  For a learnable feature $\Phi_{\bEta}(\bx) \in \RR^{m}$ with parameters $\bEta$, define the adaptive feature estimator by
  \begin{displaymath}
    f_{t}(\bx) = \Phi_{\bEta_{t}}(\bx)^{\top} \bbeta_{t},
  \end{displaymath}
  where $\bEta_{t}$ and $\bbeta_{t}$ are obtained at time $t$ by gradient flow minimizing the loss
  \begin{equation}\label{ada kernel loss}
    \cL(\bEta, \bbeta) = \left\| \bY - \Phi_{ \bEta}(\bX)^{\top} \bbeta \right\|_{2}^{2}.
  \end{equation}
\end{definition}

The feature dimension may be finite or infinite.
By analogy with \eqref{eq k equals inner of feature}, the associated time-varying kernel is
\begin{displaymath}
  k_{\bEta}(\bx,\bx^{\prime}) := \Phi_{\bEta}(\bx)^{\top} \Phi_{\bEta}(\bx^{\prime}).
\end{displaymath}
The adaptive feature model in Definition \ref{def ada kernel model} gives an abstract description of feature learning in neural networks, but the training dynamics in \eqref{ada kernel loss} remain difficult to analyze.
The main difficulties are:
\begin{itemize}
  \item [(i)] the parameterization of $\Phi_{\bEta}(\bx)$ with respect to $\bEta$ is usually complicated;
  \item [(ii)] the evolution of $\Phi_{\bEta}(\bx)$ and $\bbeta$ is coupled during training.
\end{itemize}
The neural tangent kernel (NTK) theory addresses the first difficulty by considering a sufficiently wide regime in which the feature remains approximately unchanged during training, i.e., $\Phi_{\bEta_{t}}(\bx) \approx \Phi_{\bEta_{0}}(\bx)$.
The works \cite{ba2022high}, \cite{damian2022neural}, \cite{dandi2023learning}, \cite{moniri2023theory}, and \cite{cui2024asymptotics}, discussed in Section \ref{attemps for feature learning}, simplify the second difficulty by analyzing one step of gradient descent.
\begin{revision}
\citet{lecue2025sharp} give formal definitions of alignment and feature learning, whereas \citet{gavrilopoulos2024geometrical} analyze a specified mechanism with two stages through a learned conjugate kernel and its excess risk.
The model in Definition \ref{def ada kernel model} instead serves as an abstraction of coupled feature and coefficient dynamics, and the over-parameterized Gaussian sequence model below is a simpler prototype that isolates rotation and rescaling.
\end{revision}
The following subsections develop an alternative approach to studying the dynamics in \eqref{ada kernel loss} while retaining, as far as possible, the feature learning characteristics of neural networks.
Figure \ref{figure ada_nn_kr} gives a conceptual map of the relations among adaptive feature models, neural network training, the NTK approximation, and the Gaussian sequence model prototype.

\begin{figure}[tbp]
  \centering
  {\color{black}\includegraphics[width=0.85\columnwidth]{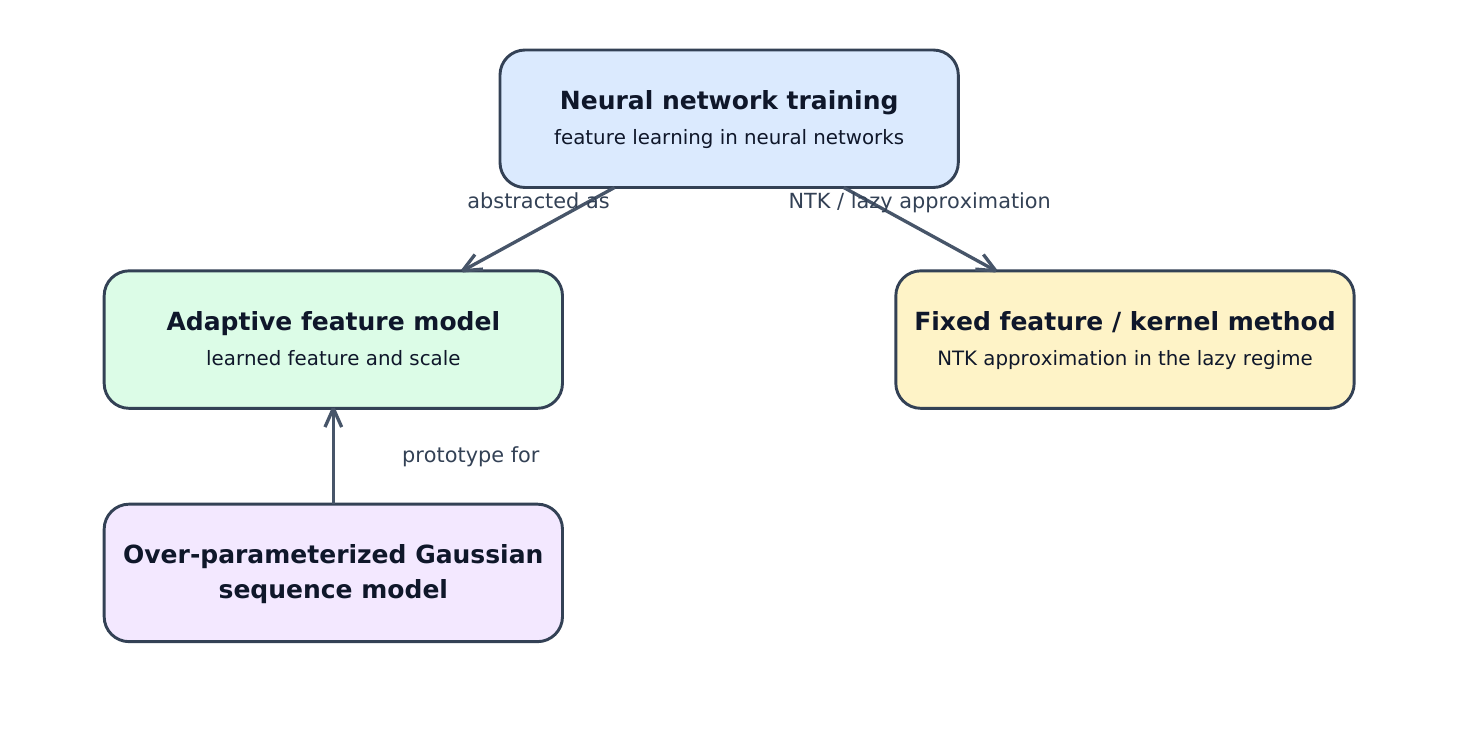}}
  {
  \caption{A conceptual map linking neural network training, the adaptive feature model, the NTK/fixed-feature approximation, and the over-parameterized Gaussian sequence model.}
  \label{figure ada_nn_kr}}
\end{figure}

We next present experiments showing that the feature learned by a neural network ($\Phi_{\bEta}(\bx)$ in \eqref{eq nn as linear in feature}) gradually aligns with the data during training.
Recall that Section \ref{subsection implications of fixed kernel} characterizes a better feature vector $\Phi(\bx) = \bLambda^{\frac{1}{2}}\Psi(\bx)$ by the concentration of the projections $\langle f^{*}, \psi_{j}\rangle_{L^{2}}$ on eigenspaces with larger eigenvalues $\lambda_{j}$.
Let $\hat{\boldsymbol{v}}_{j} \in \RR^{n}$ denote the $j$-th singular vector of the neural network feature $\Phi_{\bEta}(\bX)$, with singular values arranged in descending order.
Then $\sqrt{n}\hat{\boldsymbol{v}}_{j}$ serves as an approximation to the $j$-th eigenfunction of $\Phi_{\bEta}(\bx)$.
With $\bY \in \RR^{n}$ denoting the sample labels, we use $f_{j} := \frac{1}{\sqrt{n}}\hat{\boldsymbol{v}}_{j}^{\top}\bY$ to approximate $\langle f^{*}, \psi_{j}\rangle_{L^{2}}$.
Figure~\ref{figure ada nn exper} plots the percentage of the first $p$ projections, $\sum_{j=1}^{p} f_{j}^{2}/\sum_{j=1}^{m} f_{j}^{2}$, at each iteration $t$, for $p=1$, $100$, and $300$.
Here $m$ is the dimension of $\Phi_{\bEta}(\bx)$, and we set $m=500$ in the experiments.
The percentage increases over training, indicating improved alignment between the neural network feature and the data.
We use a fully connected two-layer neural network (FCN) on MNIST and a three-layer convolutional neural network (CNN) on CIFAR-10.
Further experimental details appear in \hyperref[section appendix experiments]{Appendix}.

\begin{figure}[tbp]
  \centering
  {\color{black}\includegraphics[width=0.45\columnwidth]{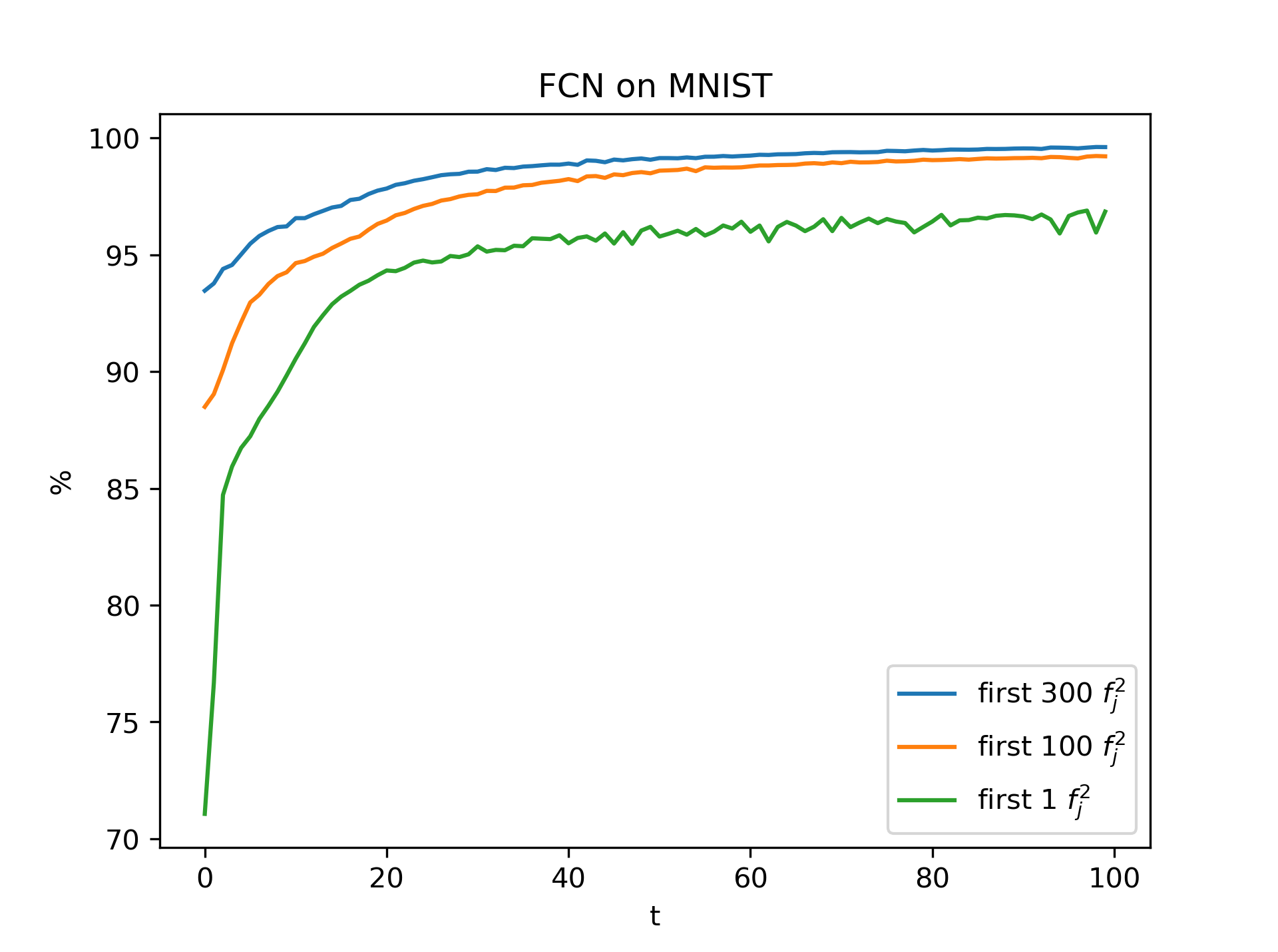}}
  {\color{black}\includegraphics[width=0.45\columnwidth]{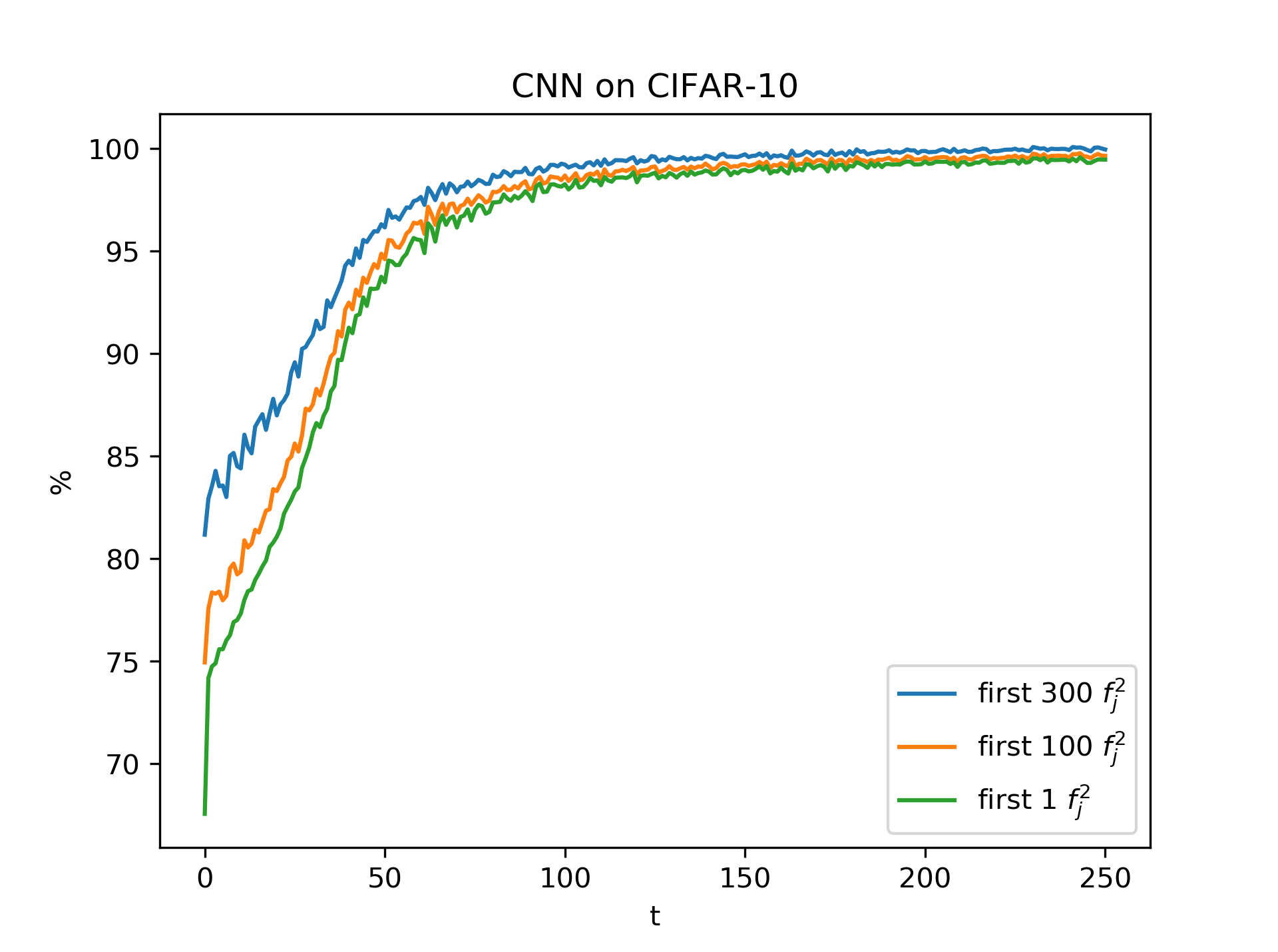}}
  {
  \caption{$x$-axis: number of training iterations; $y$-axis: the percentage of the first $p$ projections: $ \sum_{j=1}^{p} f_{j}^{2} / \sum_{j=1}^{m} f_{j}^{2}$, p = 1, 100, 300, m = 500.
  The projections $f_{j}$ concentrate on top eigenspaces as the training proceeds.}
  \label{figure ada nn exper}}
\end{figure}

\subsection{Relation between kernel regression and the Gaussian sequence model}\label{subsection relation KR GSM}

The preceding subsection introduced the adaptive feature model and the difficulty of studying its training dynamics.
Section \ref{subsection over gsm} introduces the \textit{over-parameterized Gaussian sequence model} as a prototype for the adaptive feature model.
We view the adaptive feature model as an adaptive version of kernel regression with fixed features, and the ``over-parameterized Gaussian sequence'' model as an adaptive version of the \textit{Gaussian sequence model} \citep{johnstone2017_GaussianEstimation}.

Let $k(\bx,\bx^{\prime}) = \Psi(\bx)^{\top}\bLambda\Psi(\bx^{\prime})$, where $\Psi(\bx) = (\psi_{1}(\bx), \psi_{2}(\bx), \ldots)^{\top}$ is an orthonormal basis of eigenfunctions in $L^{2}(\cX,\mu)$ and $\bLambda = \text{Diag}\{\lambda_{1}, \lambda_{2},\ldots\}$ contains the associated eigenvalues in decreasing order.
Write the target function as $f^{*}(\bx) = \sum_{j=1}^{\infty}\theta_{j}^{*}\psi_{j}(\bx)$.
Rescale the coefficients by setting $\beta_{j}^{*} = \lambda_{j}^{-\frac{1}{2}}\theta_{j}^{*}$.
Thus, $f^{*}(\bx) = \sum_{j=1}^{\infty} \lambda_{j}^{\frac{1}{2}} \beta_{j}^{*} \psi_{j}(\bx)$.
For each $j=1,2,\ldots$, let $\beta_{j,t}$ estimate $\beta_{j}^{*}$ and let $\theta_{j,t} = \lambda_{j}^{\frac{1}{2}}\beta_{j,t}$ be the corresponding estimate of $\theta_{j}^{*}$.

Kernel gradient flow with kernel $k$, equivalently linear regression in the feature space, evolves $\bbeta_{t} = (\beta_{1,t}, \beta_{2,t}, \ldots)^{\top}$ to minimize the square loss:
\begin{equation}\label{eq def dynamic kgf}
  \begin{gathered}
    \cL( \bbeta ) = \frac{1}{2n} \left\| \bY - \Psi(\bX)^{\top} \bLambda^{\frac{1}{2}} \bbeta \right\|_{2}^{2},\\ ~ \text{i.e.,}~~ \frac{\mathrm{d}}{\mathrm{d} t} \bbeta_{t} = -\nabla_{\bbeta} \cL(\bbeta_{t}),
  \end{gathered}
\end{equation}
where $\Psi(\bX) \in \RR^{\infty \times n}$, and the resulting estimator of $f^{*}$ is $\hat{f}_{t} = \Psi(\bX)^{\top}\bLambda^{\frac{1}{2}}\bbeta_{t}$.

The corresponding Gaussian sequence model observes a noisy sequence $\{z_{j}\}_{j=1}^{\infty}$ generated by:
\begin{equation}\label{def of gsm}
  z_{j} = \theta_{j}^{*} + \xi_{j}, ~j\in \NN^{+};~  \xi_{j} \overset{\text{i.i.d.}}{\sim} \mathcal{N}(0,n^{-1}),
\end{equation}
where $\{\theta_{j}^{*}\}_{j=1}^{\infty}$ are the projection coefficients $\langle f^{*}, \psi_{j}\rangle_{L^{2}}$ of $f^{*}$.
Consider the gradient flow of $\bbeta_{t} = (\beta_{1,t}, \beta_{2,t}, \ldots)^{\top}$ minimizing the square loss:
\begin{equation}\label{eq def dynamic gsm}
  \begin{gathered}
    \cL(\bbeta) = \frac{1}{2} \sum\limits_{j=1}^{\infty} \left( z_{j} - \lambda_{j}^{\frac{1}{2}} \beta_{j} \right)^{2}, \\~ \text{i.e.,}~~ \frac{\mathrm{d}}{\mathrm{d} t} \bbeta_{t} = -\nabla_{\bbeta} \cL(\bbeta_{t}),
  \end{gathered}
\end{equation}
with $\theta_{j,t} = \lambda_{j}^{\frac{1}{2}}\beta_{j,t}$ as the estimate of $\theta_{j}^{*}$.

The Gaussian sequence dynamics in this subsection are understood coordinatewise, or as limits of truncations to finitely many coordinates.
A finite truncation becomes essential in Section \ref{subsection over gsm} when we introduce a learned orthogonal transformation of the noisy coordinates.

We hypothesize a strong equivalence between kernel gradient flow \eqref{eq def dynamic kgf}, which estimates $f^{*}$, and the Gaussian sequence gradient flow \eqref{eq def dynamic gsm}, which estimates $\{\theta_{j}^{*}\}_{j=1}^{\infty}$.
We provide supporting evidence from the training dynamics and convergence rates in Sections \ref{subsubsection similar dyms} and \ref{subsubsection similar conv res}.

\subsubsection{Similar training dynamics}\label{subsubsection similar dyms}
Under gradient flow, kernel regression and the Gaussian sequence model have similar training dynamics.

For kernel gradient flow, differentiating \eqref{eq def dynamic kgf} gives
\begin{align}
  \frac{\mathrm{d}}{\mathrm{d} t}\bbeta_{t} = -\nabla_{\bbeta} \cL(\bbeta_{t}) = - \frac{1}{n} \bLambda^{\frac{1}{2}} \Psi(\bX) \Psi(\bX)^{\top} \bLambda^{\frac{1}{2}} \bbeta_{t} + \frac{1}{n} \bLambda^{\frac{1}{2}} \Psi(\bX) \bY. \notag
\end{align}
For large sample sizes, we use the rough approximations
\begin{itemize}
  \item $ \frac{1}{n} \Psi(\bX) \Psi(\bX)^{\top} \approx \mathbf{I} $;
  \item $ \left[ \frac{1}{n} \Psi(\bX) \bY \right]_{j} \approx \lambda_{j}^{\frac{1}{2}}\beta_{j}^{*} + \tilde{\epsilon}_{j}, ~~\tilde{\epsilon}_{j} \overset{\text{i.i.d.}}{\sim} \mathcal{N}(0,n^{-1})$.
\end{itemize}
Accordingly, the gradient flow of $\bbeta_{t}$ can be approximated by
\begin{equation}\label{eq dynamic kgf}
  \frac{\mathrm{d}}{\mathrm{d} t} \beta_{j,t} = -\lambda_{j} \beta_{j,t} + \lambda_{j}^{\frac{1}{2}}\left( \lambda_{j}^{\frac{1}{2}} \beta_{j}^{*} + \tilde{\epsilon}_{j}\right), \  \forall j\in \NN^+.
\end{equation}

In the Gaussian sequence model, \eqref{eq def dynamic gsm} gives, for every $j=1,2,\ldots$,
\begin{align}\label{eq dynamic gsm}
  \frac{\mathrm{d}}{\mathrm{d} t} \beta_{j,t} = -\lambda_{j} \beta_{j,t} + \lambda_{j}^{\frac{1}{2}}\left( \lambda_{j}^{\frac{1}{2}}\beta_{j}^{*}+\xi_{j}\right).
\end{align}
Because \eqref{def of gsm} assumes $\xi_{j} \overset{\text{i.i.d.}}{\sim} \mathcal{N}(0,n^{-1})$, \eqref{eq dynamic kgf} and \eqref{eq dynamic gsm} have the same form.
Thus, the two gradient flows \eqref{eq def dynamic kgf} and \eqref{eq def dynamic gsm} are approximately the same.

\subsubsection{Similar convergence rates}\label{subsubsection similar conv res}
Section \ref{subsection lc fix d} gives convergence rates for kernel gradient flow under the eigenvalue decay rate and source condition assumptions.
Under corresponding assumptions in the Gaussian sequence model, the gradient flow in \eqref{eq dynamic gsm}, or equivalently \eqref{eq def dynamic gsm}, has the same generalization error convergence rates.

With zero initialization $\beta_{j,0}=0$ for $j=1,2,\ldots$, solving \eqref{eq dynamic gsm} yields
\begin{displaymath}
  \beta_{j,t} = \lambda_{j}^{-\frac{1}{2}}(1-e^{-\lambda_j t})z_j.
\end{displaymath}
The associated estimator is $\theta_{j,t} := \lambda_{j}^{\frac{1}{2}}\beta_{j,t} = (1-e^{-\lambda_j t})z_j$.
Writing $\btheta_{t}$ and $\btheta^{*}$ for the vectors of $\theta_{j,t}$ and $\theta_{j}^{*}$, respectively, its mean squared error under \eqref{def of gsm} is
\begin{displaymath}
  \mathbb{E}\left\|\btheta_{t} - \btheta^{*} \right\|_{2}^{2} = \sum_{j=1}^\infty (e^{-\lambda_j t} \theta_j^*)^2 + \frac{1}{n} \sum_{j=1}^\infty (1-e^{-\lambda_j t})^2.
\end{displaymath}
We impose the analogous eigenvalue decay rate (EDR), $\lambda_{j} \asymp j^{-\beta}$, and source condition, $\sum_{j=1}^{\infty}\lambda_{j}^{-s}(\theta_{j}^{*})^{2}<\infty$ and $\sum_{j=1}^{\infty}\lambda_{j}^{-r}(\theta_{j}^{*})^{2}=\infty$ for every $r>s$.
Then, when $t=n^{\alpha}$ with $\alpha>0$, the calculation gives (where $\beta$ denotes the EDR)
\begin{displaymath}
  \mathbb{E}\left\|\btheta_{t} - \btheta^{*} \right\|_{2}^{2} =
  \begin{cases}
    \Theta \left(n^{- s \alpha} + n^{-(1-\alpha / \beta)}\right), & \text { if } \alpha<\beta, \\ \Omega \left(1\right), & \text { if } \alpha \geq \beta.
  \end{cases}
\end{displaymath}
These convergence rates coincide with the learning curve results for kernel gradient flow in \eqref{eq lc of kgf}.

\subsection{Over-parameterized Gaussian sequence model}\label{subsection over gsm}

\subsubsection{Applying the previous equivalence to different features/kernels}\label{subsubsection using a dif kernel}

Section \ref{subsection relation KR GSM} discusses the hypothesized equivalence between kernel regression and the Gaussian sequence model for a fixed kernel.
We now extend this equivalence to a family of kernels with different features.
To make the feature change explicit, we retain $N$ coordinates, with eigenfunctions $\Psi(\bx)=(\psi_{1}(\bx),\ldots,\psi_{N}(\bx))^{\top}$ and eigenvalue matrix $\bLambda=\mathrm{Diag}(\lambda_{1},\ldots,\lambda_{N})$.

Consider a restricted family of kernels obtained by rotating the feature coordinates and rescaling the eigenvalues relative to $\bLambda$.
For $\bA\in O(N)$ and $\bD=\mathrm{Diag}(d_{1},\ldots,d_{N})$ with $d_j>0$, define
\begin{displaymath}
  \tilde{\Psi}(\bx)=\bA^{\top}\Psi(\bx),
  \qquad
  \tilde{\bLambda}=\bD\bLambda,
  \qquad
  \tilde{\Phi}(\bx)=(\bD\bLambda)^{\frac{1}{2}}\bA^{\top}\Psi(\bx).
\end{displaymath}
The corresponding kernel is
\begin{displaymath}
  \tilde{k}(\bx,\bx^{\prime})
  =\Psi(\bx)^{\top}\bA\bD\bLambda\bA^{\top}\Psi(\bx^{\prime}).
\end{displaymath}
The truncated signal has coefficient vector $\tilde{\btheta}^{*}=\bA^{\top}\btheta^{*}$ in the new orthonormal basis.
For positive retained eigenvalues, the coefficient vector $\tilde{\bbeta}^{*}$ in the new feature representation is determined by $\tilde{\btheta}^{*}=(\bD\bLambda)^{1/2}\tilde{\bbeta}^{*}$.

Kernel gradient flow for this new feature minimizes
\begin{equation}\label{eq dynamic tilde kgf}
  \cL(\tilde{\bbeta})
  =\frac{1}{2n}\left\|\bY-\tilde{\Psi}(\bX)^{\top}(\bD\bLambda)^{\frac{1}{2}}\tilde{\bbeta}\right\|_{2}^{2}.
\end{equation}
Because $\tilde{\Psi}(\bX)=\bA^{\top}\Psi(\bX)$, the Gram matrix approximation in Section \ref{subsubsection similar dyms} is invariant under this rotation.
Under the same finite-dimensional approximation, the flow in \eqref{eq dynamic tilde kgf} has the following Gaussian sequence representation.

Let $\bZ=\btheta^{*}+\boldsymbol{\xi}$, where $\btheta^{*}\in\RR^{N}$ and $\boldsymbol{\xi}\sim\mathcal{N}(0,n^{-1}\mathbf I_N)$.
Let $\tilde{\boldsymbol{\xi}}=\bA^{\top}\boldsymbol{\xi}$ and $\tilde{\bZ}=\tilde{\btheta}^{*}+\tilde{\boldsymbol{\xi}}$.
Then $\tilde{\boldsymbol{\xi}}\sim\mathcal{N}(0,n^{-1}\mathbf I_N)$ and
\begin{equation}\label{def of gsm tilde}
  \tilde{\bZ}=\bA^{\top}\bZ.
\end{equation}
Thus, the coordinate relation in the Gaussian sequence proxy is exact, whereas the connection from kernel gradient flow to this proxy still uses the approximation from Section \ref{subsubsection similar dyms}.
The Gaussian sequence loss for the new feature is
\begin{align}
  \tilde{\cL}(\tilde{\bbeta})
  &=\frac{1}{2}\left\|\tilde{\bZ}-(\bD\bLambda)^{\frac{1}{2}}\tilde{\bbeta}\right\|_{2}^{2}
  \label{eq dynamic tilde gsm}\\
  &=\frac{1}{2}\left\|\bZ-\bA(\bD\bLambda)^{\frac{1}{2}}\tilde{\bbeta}\right\|_{2}^{2}.
  \label{eq dynamic tilde gsm trans}
\end{align}
Within this finite-dimensional family, replacing the fixed kernel or feature in kernel gradient flow corresponds to the matrix $\bA(\bD\bLambda)^{1/2}$ in the Gaussian sequence model.

\subsubsection{Over-parameterized Gaussian sequence model}

Building on this correspondence, we propose the following \textit{over-parameterized Gaussian sequence model} as a prototype for the adaptive feature model in Definition \ref{def ada kernel model}.
\begin{definition}[Over-parameterized Gaussian sequence model]\label{def over-para gsm}
  Let $\bZ=\btheta^{*}+\boldsymbol{\xi}$, where $\btheta^{*}\in\RR^{N}$ and $\boldsymbol{\xi}\sim\mathcal{N}(0,n^{-1}\mathbf I_N)$.
  For $\bA_t\in O(N)$, $\boldsymbol{s}_t\in\RR^N$, and $\boldsymbol{\alpha}_t\in\RR^N$, let
  \begin{displaymath}
    \bD_t=\mathrm{Diag}\left(e^{s_{1,t}},\ldots,e^{s_{N,t}}\right)
  \end{displaymath}
  and define
  \[
    \hat{\btheta}_t=\bA_t(\bD_t\bLambda)^{\frac{1}{2}}\boldsymbol{\alpha}_t.
  \]
  Thus, $\bD_t$ contains relative eigenvalue multipliers, while $(\bD_t\bLambda)^{1/2}$ contains the associated feature scales.
  The parameters evolve to minimize
  \begin{equation}\label{eq over gsm}
    \begin{gathered}
      \cL(\bA,\boldsymbol{s},\boldsymbol{\alpha})
      =\frac{1}{2}\left\|\bZ-\bA(\bD\bLambda)^{\frac{1}{2}}\boldsymbol{\alpha}\right\|_2^2,\\
      \frac{\mathrm d}{\mathrm dt}\bA_t=-\operatorname{grad}_{O(N)}\cL,\qquad
      \frac{\mathrm d}{\mathrm dt}\boldsymbol{s}_t=-\nabla_{\boldsymbol{s}}\cL,\qquad
      \frac{\mathrm d}{\mathrm dt}\boldsymbol{\alpha}_t=-\nabla_{\boldsymbol{\alpha}}\cL,
    \end{gathered}
  \end{equation}
  where $\operatorname{grad}_{O(N)}$ is the Riemannian gradient on the orthogonal group under the Frobenius metric.
  In this loss, $\bD=\mathrm{Diag}(e^{s_1},\ldots,e^{s_N})$.
  We initialize $\bA_0=\mathbf I_N$, $\boldsymbol{s}_0=\boldsymbol 0$, and $\boldsymbol{\alpha}_0=\boldsymbol 0$, so that $\bD_0=\mathbf I_N$ and the initial feature scale is $\bLambda^{1/2}$.
\end{definition}

The model is a prototype for the adaptive feature model: $\bA_t$ rotates feature coordinates and $\bD_t$ changes their eigenvalue scales during training.
  The contribution here is the prototype model itself, rather than a new training mechanism for neural networks or a convergence or generalization theorem for the prototype.
If $\bA_t\equiv\bA$ and $\bD_t\equiv\bD$, the loss reduces to \eqref{eq dynamic tilde gsm trans} after setting $\boldsymbol{\alpha}_t=\tilde{\bbeta}$.
The fixed feature baseline is recovered by holding $\bA_t\equiv\mathbf I_N$ and $\boldsymbol{s}_t\equiv\boldsymbol 0$.

Relative to the vanilla gradient flow in \eqref{eq def dynamic gsm}, over-parameterization refers to the extra learnable parameters $\bA_t$ and $\boldsymbol{s}_t$ in \eqref{eq over gsm}.
These parameters provide a mechanism for adapting to latent structure in $\btheta^{*}$.
They may improve generalization in suitable structures, but this is not a general guarantee and requires separate theoretical analysis.

We next describe the roles of $\bA_t$ and $\bD_t$.
Let $\boldsymbol{u}_{j,t}$ be the $j$-th column of $\bA_t$.
Along an informative trajectory, the projections $\boldsymbol{u}_{j,t}^{\top}\btheta^{*}$ align with directions having larger effective eigenvalues $d_{j,t}\lambda_j$, while directions with negligible projections have smaller effective eigenvalues.
Thus, $\bA_t$ and $\bD_t$ adjust eigenvectors and eigenvalues, respectively.

\subsubsection{Simulation}
We simulate the over-parameterized Gaussian sequence model in Definition \ref{def over-para gsm}.
Set $\boldsymbol{s}_{0}=\boldsymbol{0}$ and $\bA_{0}=\mathbf{I}_{N}$, so that $\bD_{0}=\mathbf{I}_{N}$.
To approximate \eqref{eq over gsm}, we use Riemannian gradient descent with a polar retraction after each update of $\bA_t$.
The vectors $\boldsymbol{s}_t$ and $\boldsymbol\alpha_t$ are updated by ordinary gradient descent.
We use $N=500$, $n=4000$, and a learning rate of $0.5$.
The signal and eigenvalues are $\theta_{j}^{*}=1/(N-j+2)$ and $\lambda_{j}=1/(j+5)^{2}$ for $j=1,\ldots,N$.
Thus, $\{\lambda_{j}\}_{j=1}^{N}$ decreases while $\{\theta_{j}^{*}\}_{j=1}^{N}$ increases.

For each $j=1,2,\ldots,N$, let $j^{\prime}$ index the $j$-th largest effective eigenvalue $d_{j^{\prime},t}\lambda_{j^{\prime}}$, and let $\boldsymbol{u}_{j^{\prime},t}$ be the corresponding column of $\bA_t$.
Let $f_{j,t} := \boldsymbol{u}_{j^{\prime},t}^{\top} \btheta^{*}$.
Then $f_{j,t}$ is the projection onto the eigenvector corresponding to the $j$-th largest effective eigenvalue.

At initialization, $\bA_{0}=\mathbf{I}_{N}$, $\bD_{0}=\mathbf I_N$, and $f_{j,0}=\theta_{j}^{*}$, so the data are misaligned with the feature: the coefficients $\theta_{j}^{*}$ are smaller in directions with larger $\lambda_{j}$.
The quantities $f_{j,t}$ are oracle diagnostics in this simulation because they use the known vector $\btheta^{*}$.
Figure \ref{figure over gsm} shows that the percentage of the first $p$ projections, $\sum_{j=1}^{p} f_{j,t}^{2}/\sum_{j=1}^{N} f_{j,t}^{2}$ for $p=100$ and $300$, increases with the number of training iterations.

\begin{figure}[tbp]
  \centering
  {\color{black}\includegraphics[width=0.45\columnwidth]{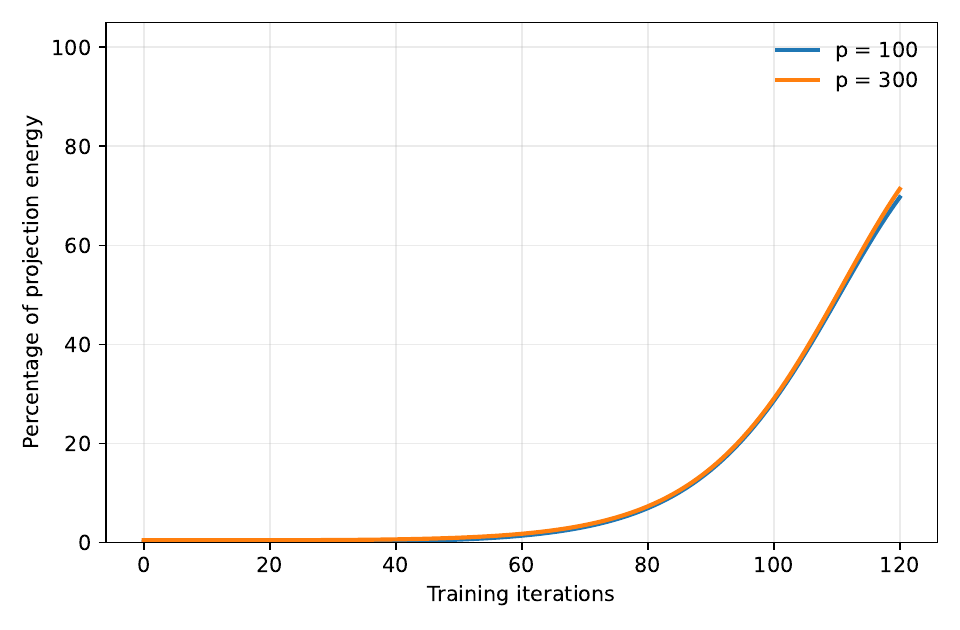}}
  {
  \caption{$x$-axis: number of training iterations; $y$-axis: the percentage of the first $p$ projections: $ \sum_{j=1}^{p} f_{j,t}^{2} / \sum_{j=1}^{N} f_{j,t}^{2}$, $p = 100, 300, N = 500$.
  The projection energy of $f_{j,t}$ increasingly concentrates on directions with the largest effective eigenvalues during training.}
  \label{figure over gsm}}
\end{figure}

\subsection{Over-parameterized models in related literature}

\cite{li2024improving} studied a data-generating model similar to that in Definition \ref{def over-para gsm}:
\begin{displaymath}
  z_{j} = \theta_{j}^{*} + \xi_{j}, ~~j = 1,2, \ldots,
\end{displaymath}
where the noise variables $\{\xi_{j}\}_{j=1}^{\infty}$ are $\epsilon^{2}$-sub-Gaussian.
They estimated $\theta_{j}^{*}$ by $\theta_{j,t}:=a_{j,t}\beta_{j,t}$ and applied gradient flow to $\boldsymbol{a}_{t}=(a_{1,t}, a_{2,t}, \ldots)^{\top}$ and $\bbeta_{t}=(\beta_{1,t}, \beta_{2,t}, \ldots)^{\top}$ to minimize
\begin{equation}\label{eq dym of exm over}
  \cL(\boldsymbol{a}, \bbeta) = \frac{1}{2} \sum\limits_{j=1}^{\infty} \left(z_{j} - a_{j} \beta_{j} \right)^{2}.
\end{equation}
They initialized $a_{j,0}=\lambda_{j}^{\frac{1}{2}}$ for a sequence $\lambda_{j}\asymp j^{-\beta}$ with $\beta>1$ and set $\beta_{j,0}=0$ for $j=1,2,\ldots$.
Let $\btheta_{t}$ and $\btheta^{*}$ denote the vectors of $\theta_{j,t}$ and $\theta_{j}^{*}$, respectively.

They also compared the generalization error of the gradient flow \eqref{eq dym of exm over} with that of the vanilla gradient flow \eqref{eq def dynamic gsm}, in which $\boldsymbol{a}_{t}$ remains fixed at $a_{j,t}\equiv\lambda_{j}^{\frac{1}{2}}$ for $j=1,2,\ldots$.
Corollary 3.3 of \cite{li2024improving} provides a concrete example where the gradient flow \eqref{eq dym of exm over} has a faster generalization error convergence rate than the vanilla gradient flow \eqref{eq def dynamic gsm}.

For a finite truncation with $a_{j,t}\geq 0$ and $\lambda_j>0$, setting $\bA_t\equiv\mathbf I_N$ and choosing $d_{j,t}=a_{j,t}^{2}/\lambda_j$ gives $(\bD_t\bLambda)^{1/2}=\mathrm{Diag}\{a_{1,t},\ldots,a_{N,t}\}$.
Thus, \eqref{eq dym of exm over} is the counterpart of Definition \ref{def over-para gsm} with only diagonal rescaling, whereas our prototype also permits rotations of the feature coordinates.
The rotational extension may improve generalization in suitable structures, but a general comparison remains an open theoretical question.

A separate line of research studies the advantages of over-parameterization in high-dimensional linear regression with sparse signals \citep{vaskevicius2019implicit,li2021implicit,zhao2022high}.
For example, \cite{zhao2022high} applied gradient descent to $\boldsymbol{g}_{t}, \boldsymbol{l}_{t}\in\RR^{d}$ to minimize the square loss
\begin{displaymath}
  \cL(\boldsymbol{g}, \boldsymbol{l}) = \frac{1}{2 n}\| \bY - \bX^{\top}(\boldsymbol{g} \circ \boldsymbol{l}) \|_{2}^{2},
\end{displaymath}
where ``$\circ$'' denotes the Hadamard product.
They showed that the over-parameterized version of gradient descent could lead to a near-minimax optimal rate and a faster dimension-free convergence rate in the strong signal case.
\end{revision}
 \section{Discussion}\label{section discussion}

Theoretically analyzing the feature learning characteristics of neural network models is a challenging task.
The primary difficulty lies in the complex training dynamics of these models.
Although existing results have made valuable progress (Section \ref{attemps for feature learning}), fully understanding feature learning and its benefits for generalization ability still requires considerable effort.
\begin{revision}
This paper introduces the over-parameterized Gaussian sequence model (Definition \ref{def over-para gsm}) as a prototype for the adaptive feature model.
This prototype borrows the concept of a good feature from fixed kernel regression theory and retains the rotation and rescaling mechanisms of adaptive feature learners.
Relative to \citet{lecue2025sharp} and \citet{gavrilopoulos2024geometrical}, the contribution is this prototype model rather than a new general definition of feature learning, a training mechanism for neural networks, or a mathematical guarantee.
\end{revision}

\begin{revision}
We expect that the over-parameterized Gaussian sequence model can offer insights into how feature learning occurs and may improve generalization in suitable structures.
As a prototype, while it does not by itself transfer results to real neural networks, we expect to establish such a connection in the future.
\end{revision}
To this end, we need to investigate (but not limited to) the following questions in the future:
\begin{itemize}[leftmargin=*, itemindent=12pt]
\item Theoretical understanding of the over-parameterized Gaussian sequence model in Definition \ref{def over-para gsm}.
Specifically, the first step is to characterize the alignment between $\bA_{t}, \bD_{t}$ and the true parameters as the gradient flow progresses.
The next step is to analyze whether the alignment improves generalization ability relative to vanilla gradient flow or models without over-parameterization.
We can begin with a simple family of feature maps and gradually increase their complexity.
For example, \cite{li2024improving} considered the simplest case where the eigenvalues are adjustable.
In addition, one can consider the case where $\bA_{t}$ varies among a family of orthogonal matrices, making the eigenvectors adjustable.
\item Rigorous equivalence between the (adaptive feature) kernel regression model and the (over-parameterized) Gaussian sequence model, especially in the high dimensions.
We have shown some evidence of the equivalence in Section \ref{subsubsection similar conv res}, but the rigorous proof and the high-dimensional case remain open problems.
Establishing this equivalence is required before conclusions from the prototype can be extended to the adaptive feature model or real neural networks.
\item The specific architecture of neural networks (e.g., the number of layers, fully connected or convolutional neural networks, etc.) will inevitably affect the manner and efficiency of feature learning.
Characterizing the differences between various neural network architectures using the adaptive feature model will be a more advanced problem.
\end{itemize}
 \section*{Experimental details}\label{section appendix experiments}

We provide some details of the neural networks experiments in Section \ref{subsection nn as ada kernel}.
\begin{itemize}
\item The fully connected neural network (for MNIST).
Input dimension = 784; first hidden layer dimension = 500; second hidden layer dimension = 500; readout layer dimension = 1.
\item The convolutional neural network (for CIFAR-10).
Input shape = $3*32*32$; first convolutional layer: 64 channels, $5*5$ kernel, dropout = 0.5, max pooling; second convolutional layer: 256 channels, $5*5$ kernel, dropout = 0.5, average pooling; third convolutional layer: 32 channels, $5*5$ kernel, dropout = 0.5, no pooling; a fully connected layer: dimension=500; readout layer: dimension = 1.
\end{itemize}

We use ReLU activation, MSE loss, batch size of 32 and a learning rate of 0.001.
We train neural networks in all 60000 samples of MNIST or CIFAR-10 and use 10000 of these samples to calculate $f_{j} := \frac{1}{\sqrt{n}} \hat{\boldsymbol{v}}_{j}^{\top} \bY$ (i.e., $n=10000$) in Section \ref{subsection nn as ada kernel}.
 \section*{Acknowledgments}
Qian Lin was supported in part by the National Natural Science Foundation of China (Grant 92370122, Grant 11971257).


\begin{thebibliography}{132}
\providecommand{\natexlab}[1]{#1}
\providecommand{\url}[1]{\texttt{#1}}
\expandafter\ifx\csname urlstyle\endcsname\relax
  \providecommand{\doi}[1]{doi: #1}\else
  \providecommand{\doi}{doi: \begingroup \urlstyle{rm}\Url}\fi

\bibitem[Abbe et~al.(2023)Abbe, Boix-Adsera, and Misiakiewicz]{abbe2023leap}
Emmanuel Abbe, Enric Boix-Adsera, and Theodor Misiakiewicz.
\newblock {SGD} learning on neural networks: Leap complexity and
  saddle-to-saddle dynamics.
\newblock \emph{arXiv preprint arXiv:2302.11055}, 2023.
\newblock \doi{10.48550/arXiv.2302.11055}.
\newblock URL \url{https://arxiv.org/abs/2302.11055}.

\bibitem[Abbe et~al.(2024)Abbe, Boix-Adsera, and Misiakiewicz]{abbe2024merged}
Emmanuel Abbe, Enric Boix-Adsera, and Theodor Misiakiewicz.
\newblock The merged-staircase property: A necessary and nearly sufficient
  condition for {SGD} learning of sparse functions on two-layer neural
  networks.
\newblock \emph{arXiv preprint arXiv:2202.08658}, 2024.
\newblock \doi{10.48550/arXiv.2202.08658}.
\newblock URL \url{https://arxiv.org/abs/2202.08658}.

\bibitem[Adams and Fournier(2003)]{adams2003_SobolevSpaces}
Robert~A Adams and John~JF Fournier.
\newblock \emph{Sobolev Spaces}.
\newblock {Elsevier}, 2003.

\bibitem[Aerni et~al.(2022)Aerni, Milanta, Donhauser, and
  Yang]{aerni2022strong}
Michael Aerni, Marco Milanta, Konstantin Donhauser, and Fanny Yang.
\newblock Strong inductive biases provably prevent harmless interpolation.
\newblock In \emph{The Eleventh International Conference on Learning
  Representations}, 2022.

\bibitem[Arora et~al.(2019{\natexlab{a}})Arora, Du, Hu, Li, and
  Wang]{arora2019fine}
Sanjeev Arora, Simon Du, Wei Hu, Zhiyuan Li, and Ruosong Wang.
\newblock Fine-grained analysis of optimization and generalization for
  overparameterized two-layer neural networks.
\newblock In \emph{International Conference on Machine Learning}, pages
  322--332. PMLR, 2019{\natexlab{a}}.

\bibitem[Arora et~al.(2019{\natexlab{b}})Arora, Du, Hu, Li, Salakhutdinov, and
  Wang]{arora2019exact}
Sanjeev Arora, Simon~S Du, Wei Hu, Zhiyuan Li, Russ~R Salakhutdinov, and
  Ruosong Wang.
\newblock On exact computation with an infinitely wide neural net.
\newblock \emph{Advances in neural information processing systems}, 32,
  2019{\natexlab{b}}.

\bibitem[Atanasov et~al.(2021)Atanasov, Bordelon, and
  Pehlevan]{atanasov2021neural}
Alexander Atanasov, Blake Bordelon, and Cengiz Pehlevan.
\newblock Neural networks as kernel learners: The silent alignment effect.
\newblock In \emph{International Conference on Learning Representations}, 2021.

\bibitem[Ba et~al.(2022)Ba, Erdogdu, Suzuki, Wang, Wu, and Yang]{ba2022high}
Jimmy Ba, Murat~A Erdogdu, Taiji Suzuki, Zhichao Wang, Denny Wu, and Greg Yang.
\newblock High-dimensional asymptotics of feature learning: How one gradient
  step improves the representation.
\newblock \emph{Advances in Neural Information Processing Systems},
  35:\penalty0 37932--37946, 2022.

\bibitem[Baratin et~al.(2021)Baratin, George, Laurent, Hjelm, Lajoie, Vincent,
  and Lacoste-Julien]{baratin2021implicit}
Aristide Baratin, Thomas George, C{\'e}sar Laurent, R~Devon Hjelm, Guillaume
  Lajoie, Pascal Vincent, and Simon Lacoste-Julien.
\newblock Implicit regularization via neural feature alignment.
\newblock In \emph{International Conference on Artificial Intelligence and
  Statistics}, pages 2269--2277. PMLR, 2021.

\bibitem[Bartlett et~al.(2020)Bartlett, Long, Lugosi, and
  Tsigler]{bartlett2020benign}
Peter~L Bartlett, Philip~M Long, G{\'a}bor Lugosi, and Alexander Tsigler.
\newblock Benign overfitting in linear regression.
\newblock \emph{Proceedings of the National Academy of Sciences}, 117\penalty0
  (48):\penalty0 30063--30070, 2020.

\bibitem[Bartlett et~al.(2021)Bartlett, Montanari, and
  Rakhlin]{bartlett2021deep}
Peter~L Bartlett, Andrea Montanari, and Alexander Rakhlin.
\newblock Deep learning: a statistical viewpoint.
\newblock \emph{Acta numerica}, 30:\penalty0 87--201, 2021.

\bibitem[Barzilai and Shamir(2023)]{barzilai2023generalization}
Daniel Barzilai and Ohad Shamir.
\newblock Generalization in kernel regression under realistic assumptions.
\newblock \emph{arXiv preprint arXiv:2312.15995}, 2023.

\bibitem[Bauer and Kohler(2019)]{bauer2019deep}
Benedikt Bauer and Michael Kohler.
\newblock On deep learning as a remedy for the curse of dimensionality in
  nonparametric regression.
\newblock \emph{The Annals of Statistics}, 47\penalty0 (4):\penalty0
  2261--2285, 2019.

\bibitem[Bauer et~al.(2007)Bauer, Pereverzyev, and
  Rosasco]{bauer2007_RegularizationAlgorithms}
F.~Bauer, S.~Pereverzyev, and L.~Rosasco.
\newblock On regularization algorithms in learning theory.
\newblock \emph{Journal of complexity}, 23\penalty0 (1):\penalty0 52--72, 2007.

\bibitem[Beaglehole et~al.(2023{\natexlab{a}})Beaglehole, Belkin, and
  Pandit]{beaglehole2023inconsistency}
Daniel Beaglehole, Mikhail Belkin, and Parthe Pandit.
\newblock On the inconsistency of kernel ridgeless regression in fixed
  dimensions.
\newblock \emph{SIAM Journal on Mathematics of Data Science}, 5\penalty0
  (4):\penalty0 854--872, 2023{\natexlab{a}}.

\bibitem[Beaglehole et~al.(2023{\natexlab{b}})Beaglehole, Radhakrishnan,
  Pandit, and Belkin]{beaglehole2023mechanism}
Daniel Beaglehole, Adityanarayanan Radhakrishnan, Parthe Pandit, and Mikhail
  Belkin.
\newblock Mechanism of feature learning in convolutional neural networks.
\newblock \emph{arXiv preprint arXiv:2309.00570}, 2023{\natexlab{b}}.
\newblock \doi{10.48550/arXiv.2309.00570}.
\newblock URL \url{https://arxiv.org/abs/2309.00570}.

\bibitem[Beaglehole et~al.(2024)Beaglehole, Mitliagkas, and
  Agarwala]{beaglehole2024gradient}
Daniel Beaglehole, Ioannis Mitliagkas, and Atish Agarwala.
\newblock Gradient descent induces alignment between weights and the empirical
  ntk for deep non-linear networks.
\newblock \emph{arXiv preprint arXiv:2402.05271}, 2024.

\bibitem[Belkin(2021)]{belkin2021fit}
Mikhail Belkin.
\newblock Fit without fear: remarkable mathematical phenomena of deep learning
  through the prism of interpolation.
\newblock \emph{Acta Numerica}, 30:\penalty0 203--248, 2021.

\bibitem[Belkin et~al.(2018)Belkin, Ma, and Mandal]{Belkin2018ToUD}
Mikhail Belkin, Siyuan Ma, and Soumik Mandal.
\newblock To understand deep learning we need to understand kernel learning.
\newblock In \emph{International Conference on Machine Learning}, 2018.
\newblock URL \url{https://api.semanticscholar.org/CorpusID:3617641}.

\bibitem[Ben~Arous et~al.(2021)Ben~Arous, Gheissari, and
  Jagannath]{arous2021online}
Gerard Ben~Arous, Reza Gheissari, and Aukosh Jagannath.
\newblock Online stochastic gradient descent on non-convex losses from
  high-dimensional inference.
\newblock \emph{Journal of Machine Learning Research}, 22\penalty0
  (106):\penalty0 1--51, 2021.
\newblock URL \url{https://jmlr.org/papers/v22/20-1288.html}.

\bibitem[Blanchard and M{\"u}cke(2018)]{blanchard2018_OptimalRates}
G.~Blanchard and Nicole M{\"u}cke.
\newblock Optimal rates for regularization of statistical inverse learning
  problems.
\newblock \emph{Foundations of Computational Mathematics}, 18:\penalty0
  971--1013, 2018.

\bibitem[Bordelon and Pehlevan(2022)]{bordelon2022self}
Blake Bordelon and Cengiz Pehlevan.
\newblock Self-consistent dynamical field theory of kernel evolution in wide
  neural networks.
\newblock \emph{Advances in Neural Information Processing Systems},
  35:\penalty0 32240--32256, 2022.

\bibitem[Bordelon and Pehlevan(2024)]{bordelon2024dynamics}
Blake Bordelon and Cengiz Pehlevan.
\newblock Dynamics of finite width kernel and prediction fluctuations in mean
  field neural networks.
\newblock \emph{Advances in Neural Information Processing Systems}, 36, 2024.

\bibitem[Bordelon et~al.(2020)Bordelon, Canatar, and
  Pehlevan]{Bordelon2020SpectrumDL}
Blake Bordelon, Abdulkadir Canatar, and Cengiz Pehlevan.
\newblock Spectrum dependent learning curves in kernel regression and wide
  neural networks.
\newblock In \emph{ICML}, 2020.

\bibitem[Buchholz(2022)]{buchholz2022kernel}
Simon Buchholz.
\newblock Kernel interpolation in sobolev spaces is not consistent in low
  dimensions.
\newblock In \emph{Conference on Learning Theory}, pages 3410--3440. PMLR,
  2022.

\bibitem[Caponnetto(2006)]{caponnetto2006optimal}
Andrea Caponnetto.
\newblock Optimal rates for regularization operators in learning theory.
\newblock Technical report, MASSACHUSETTS INST OF TECH CAMBRIDGE COMPUTER
  SCIENCE AND ARTIFICIAL~…, 2006.

\bibitem[Caponnetto and de~Vito(2007)]{Caponnetto2007OptimalRF}
Andrea Caponnetto and Ernesto de~Vito.
\newblock Optimal rates for the regularized least-squares algorithm.
\newblock \emph{Foundations of Computational Mathematics}, 7:\penalty0
  331--368, 2007.

\bibitem[Celisse and Wahl(2020)]{Celisse2020AnalyzingTD}
Alain Celisse and Martin Wahl.
\newblock Analyzing the discrepancy principle for kernelized spectral filter
  learning algorithms.
\newblock \emph{J. Mach. Learn. Res.}, 22:\penalty0 76:1--76:59, 2020.

\bibitem[Chen et~al.(2024)Chen, Li, and Lin]{Chen2024OnTI}
Guhan Chen, Yicheng Li, and Qian Lin.
\newblock On the impacts of the random initialization in the neural tangent
  kernel theory.
\newblock \emph{arXiv preprint arXiv:2410.05626}, 2024.
\newblock URL \url{https://arxiv.org/pdf/2410.05626}.

\bibitem[Cheng et~al.(2024)Cheng, Lucchi, Kratsios, and
  Belius]{cheng2024characterizing}
Tin~Sum Cheng, Aurelien Lucchi, Anastasis Kratsios, and David Belius.
\newblock Characterizing overfitting in kernel ridgeless regression through the
  eigenspectrum.
\newblock \emph{arXiv preprint arXiv:2402.01297}, 2024.

\bibitem[Chizat et~al.(2019)Chizat, Oyallon, and Bach]{chizat2019lazy}
Lenaic Chizat, Edouard Oyallon, and Francis Bach.
\newblock On lazy training in differentiable programming.
\newblock \emph{Advances in neural information processing systems},
  32:\penalty0 2937--2947, 2019.
\newblock URL
  \url{https://proceedings.neurips.cc/paper_files/paper/2019/hash/ae614c557843b1df326cb29c57225459-Abstract.html}.

\bibitem[Cortes et~al.(2012)Cortes, Mohri, and
  Rostamizadeh]{cortes2012algorithms}
Corinna Cortes, Mehryar Mohri, and Afshin Rostamizadeh.
\newblock Algorithms for learning kernels based on centered alignment.
\newblock \emph{The Journal of Machine Learning Research}, 13\penalty0
  (1):\penalty0 795--828, 2012.

\bibitem[Cucker and Smale(2001)]{Cucker2001OnTM}
Felipe Cucker and Stephen Smale.
\newblock On the mathematical foundations of learning.
\newblock \emph{Bulletin of the American Mathematical Society}, 39:\penalty0
  1--49, 2001.
\newblock URL \url{https://api.semanticscholar.org/CorpusID:8188805}.

\bibitem[Cui et~al.(2021)Cui, Loureiro, Krzakala, and
  Zdeborov{\'a}]{Cui2021GeneralizationER}
Hugo Cui, Bruno Loureiro, Florent Krzakala, and Lenka Zdeborov{\'a}.
\newblock Generalization error rates in kernel regression: The crossover from
  the noiseless to noisy regime.
\newblock \emph{Advances in Neural Information Processing Systems},
  34:\penalty0 10131--10143, 2021.

\bibitem[Cui et~al.(2024)Cui, Pesce, Dandi, Krzakala, Lu, Zdeborov{\'a}, and
  Loureiro]{cui2024asymptotics}
Hugo Cui, Luca Pesce, Yatin Dandi, Florent Krzakala, Yue~M Lu, Lenka
  Zdeborov{\'a}, and Bruno Loureiro.
\newblock Asymptotics of feature learning in two-layer networks after one
  gradient-step.
\newblock \emph{arXiv preprint arXiv:2402.04980}, 2024.

\bibitem[Damian et~al.(2022)Damian, Lee, and Soltanolkotabi]{damian2022neural}
Alexandru Damian, Jason Lee, and Mahdi Soltanolkotabi.
\newblock Neural networks can learn representations with gradient descent.
\newblock In \emph{Conference on Learning Theory}, pages 5413--5452. PMLR,
  2022.

\bibitem[Dandi et~al.(2023)Dandi, Krzakala, Loureiro, Pesce, and
  Stephan]{dandi2023learning}
Yatin Dandi, Florent Krzakala, Bruno Loureiro, Luca Pesce, and Ludovic Stephan.
\newblock Learning two-layer neural networks, one (giant) step at a time.
\newblock \emph{arXiv preprint arXiv:2305.18270}, 2023.

\bibitem[Dieuleveut and Bach(2016)]{dieuleveut2016nonparametric}
Aymeric Dieuleveut and Francis Bach.
\newblock Nonparametric stochastic approximation with large step-sizes1.
\newblock \emph{THE ANNALS}, 44\penalty0 (4):\penalty0 1363--1399, 2016.

\bibitem[Donhauser et~al.(2021)Donhauser, Wu, and Yang]{Donhauser_how_2021}
Konstantin Donhauser, Mingqi Wu, and Fanny Yang.
\newblock How rotational invariance of common kernels prevents generalization
  in high dimensions.
\newblock In \emph{International Conference on Machine Learning}, pages
  2804--2814. PMLR, 2021.

\bibitem[Du et~al.(2019)Du, Lee, Li, Wang, and Zhai]{du2019gradient}
Simon Du, Jason Lee, Haochuan Li, Liwei Wang, and Xiyu Zhai.
\newblock Gradient descent finds global minima of deep neural networks.
\newblock In \emph{International conference on machine learning}, pages
  1675--1685. PMLR, 2019.

\bibitem[Dyer and Gur-Ari(2019)]{dyer2019asymptotics}
Ethan Dyer and Guy Gur-Ari.
\newblock Asymptotics of wide networks from feynman diagrams.
\newblock In \emph{International Conference on Learning Representations}, 2019.

\bibitem[Edmunds and Triebel(1996)]{edmunds_triebel_1996}
D.~E. Edmunds and H.~Triebel.
\newblock \emph{Function Spaces, Entropy Numbers, Differential Operators}.
\newblock Cambridge Tracts in Mathematics. Cambridge University Press, 1996.
\newblock \doi{10.1017/CBO9780511662201}.

\bibitem[Fan et~al.(2021)Fan, Ma, and Zhong]{fan2021selective}
Jianqing Fan, Cong Ma, and Yiqiao Zhong.
\newblock A selective overview of deep learning.
\newblock \emph{Statistical science: a review journal of the Institute of
  Mathematical Statistics}, 36\penalty0 (2):\penalty0 264, 2021.

\bibitem[Fischer and Steinwart(2020)]{fischer2020_SobolevNorm}
Simon-Raphael Fischer and Ingo Steinwart.
\newblock Sobolev norm learning rates for regularized least-squares algorithms.
\newblock \emph{Journal of Machine Learning Research}, 21:\penalty0
  205:1--205:38, 2020.

\bibitem[Fort et~al.(2020)Fort, Dziugaite, Paul, Kharaghani, Roy, and
  Ganguli]{fort2020deep}
Stanislav Fort, Gintare~Karolina Dziugaite, Mansheej Paul, Sepideh Kharaghani,
  Daniel~M Roy, and Surya Ganguli.
\newblock Deep learning versus kernel learning: an empirical study of loss
  landscape geometry and the time evolution of the neural tangent kernel.
\newblock \emph{Advances in Neural Information Processing Systems},
  33:\penalty0 5850--5861, 2020.

\bibitem[Gavrilopoulos et~al.(2024)Gavrilopoulos, Lecu{\'e}, and
  Shang]{gavrilopoulos2024geometrical}
Georgios Gavrilopoulos, Guillaume Lecu{\'e}, and Zong Shang.
\newblock A geometrical analysis of kernel ridge regression and its
  applications.
\newblock \emph{arXiv preprint arXiv:2404.07709}, 2024.

\bibitem[Geiger et~al.(2020)Geiger, Spigler, Jacot, and
  Wyart]{geiger2020disentangling}
Mario Geiger, Stefano Spigler, Arthur Jacot, and Matthieu Wyart.
\newblock Disentangling feature and lazy training in deep neural networks.
\newblock \emph{Journal of Statistical Mechanics: Theory and Experiment},
  2020\penalty0 (11):\penalty0 113301, 2020.

\bibitem[Gerace et~al.(2020)Gerace, Loureiro, Krzakala, M{\'e}zard, and
  Zdeborov{\'a}]{gerace2020generalisation}
Federica Gerace, Bruno Loureiro, Florent Krzakala, Marc M{\'e}zard, and Lenka
  Zdeborov{\'a}.
\newblock Generalisation error in learning with random features and the hidden
  manifold model.
\newblock In \emph{International Conference on Machine Learning}, pages
  3452--3462. PMLR, 2020.

\bibitem[Gerfo et~al.(2008)Gerfo, Rosasco, Odone, Vito, and
  Verri]{gerfo2008_SpectralAlgorithms}
L.~Lo Gerfo, Lorenzo Rosasco, Francesca Odone, E.~De Vito, and Alessandro
  Verri.
\newblock Spectral algorithms for supervised learning.
\newblock \emph{Neural Computation}, 20\penalty0 (7):\penalty0 1873--1897,
  2008.

\bibitem[Ghorbani et~al.(2019)Ghorbani, Mei, Misiakiewicz, and
  Montanari]{ghorbani2019limitations}
Behrooz Ghorbani, Song Mei, Theodor Misiakiewicz, and Andrea Montanari.
\newblock Limitations of lazy training of two-layers neural network.
\newblock \emph{Advances in Neural Information Processing Systems}, 32, 2019.

\bibitem[Ghorbani et~al.(2020)Ghorbani, Mei, Misiakiewicz, and
  Montanari]{Ghorbani_When_2021}
Behrooz Ghorbani, Song Mei, Theodor Misiakiewicz, and Andrea Montanari.
\newblock When do neural networks outperform kernel methods?
\newblock \emph{Advances in Neural Information Processing Systems},
  33:\penalty0 14820--14830, 2020.

\bibitem[Ghorbani et~al.(2021)Ghorbani, Mei, Misiakiewicz, and
  Montanari]{Ghorbani2019LinearizedTN}
Behrooz Ghorbani, Song Mei, Theodor Misiakiewicz, and Andrea Montanari.
\newblock {Linearized two-layers neural networks in high dimension}.
\newblock \emph{The Annals of Statistics}, 49\penalty0 (2):\penalty0 1029 --
  1054, 2021.
\newblock \doi{10.1214/20-AOS1990}.
\newblock URL \url{https://doi.org/10.1214/20-AOS1990}.

\bibitem[Ghosh et~al.(2021)Ghosh, Mei, and Yu]{Ghosh_three_2021}
Nikhil Ghosh, Song Mei, and Bin Yu.
\newblock The three stages of learning dynamics in high-dimensional kernel
  methods.
\newblock In \emph{International Conference on Learning Representations}, 2021.

\bibitem[Golikov et~al.(2022)Golikov, Pokonechnyy, and
  Korviakov]{golikov2022ntk}
Eugene Golikov, Eduard Pokonechnyy, and Vladimir Korviakov.
\newblock Neural tangent kernel: A survey.
\newblock \emph{arXiv preprint arXiv:2208.13614}, 2022.
\newblock \doi{10.48550/arXiv.2208.13614}.
\newblock URL \url{https://arxiv.org/abs/2208.13614}.

\bibitem[Haas et~al.(2024)Haas, Holzm{\"u}ller, Luxburg, and
  Steinwart]{haas2024mind}
Moritz Haas, David Holzm{\"u}ller, Ulrike Luxburg, and Ingo Steinwart.
\newblock Mind the spikes: Benign overfitting of kernels and neural networks in
  fixed dimension.
\newblock \emph{Advances in Neural Information Processing Systems}, 36, 2024.

\bibitem[Hanin and Nica(2019)]{hanin2019finite}
Boris Hanin and Mihai Nica.
\newblock Finite depth and width corrections to the neural tangent kernel.
\newblock \emph{arXiv preprint arXiv:1909.05989}, 2019.

\bibitem[Hastie et~al.(2022)Hastie, Montanari, Rosset, and
  Tibshirani]{10.1214/21-AOS2133}
Trevor Hastie, Andrea Montanari, Saharon Rosset, and Ryan~J. Tibshirani.
\newblock {Surprises in high-dimensional ridgeless least squares
  interpolation}.
\newblock \emph{The Annals of Statistics}, 50\penalty0 (2):\penalty0 949 --
  986, 2022.
\newblock \doi{10.1214/21-AOS2133}.
\newblock URL \url{https://doi.org/10.1214/21-AOS2133}.

\bibitem[Hu and Lu(2022{\natexlab{a}})]{hu2022sharp}
Hong Hu and Yue~M Lu.
\newblock Sharp asymptotics of kernel ridge regression beyond the linear
  regime.
\newblock \emph{arXiv preprint arXiv:2205.06798}, 2022{\natexlab{a}}.

\bibitem[Hu and Lu(2022{\natexlab{b}})]{hu2022universality}
Hong Hu and Yue~M Lu.
\newblock Universality laws for high-dimensional learning with random features.
\newblock \emph{IEEE Transactions on Information Theory}, 69\penalty0
  (3):\penalty0 1932--1964, 2022{\natexlab{b}}.

\bibitem[Hu et~al.(2019)Hu, Li, and Yu]{hu2019simple}
Wei Hu, Zhiyuan Li, and Dingli Yu.
\newblock Simple and effective regularization methods for training on noisily
  labeled data with generalization guarantee.
\newblock \emph{arXiv preprint arXiv:1905.11368}, 2019.

\bibitem[Huang and Yau(2020)]{huang2020dynamics}
Jiaoyang Huang and Horng-Tzer Yau.
\newblock Dynamics of deep neural networks and neural tangent hierarchy.
\newblock In \emph{International conference on machine learning}, pages
  4542--4551. PMLR, 2020.

\bibitem[Huang et~al.(2020)Huang, Wang, Tao, and Zhao]{huang2020deep}
Kaixuan Huang, Yuqing Wang, Molei Tao, and Tuo Zhao.
\newblock Why do deep residual networks generalize better than deep feedforward
  networks?---a neural tangent kernel perspective.
\newblock \emph{Advances in neural information processing systems},
  33:\penalty0 2698--2709, 2020.

\bibitem[Jacot et~al.(2018)Jacot, Gabriel, and Hongler]{jacot2018neural}
Arthur Jacot, Franck Gabriel, and Cl{\'e}ment Hongler.
\newblock Neural tangent kernel: Convergence and generalization in neural
  networks.
\newblock \emph{Advances in neural information processing systems}, 31, 2018.

\bibitem[Jiang et~al.(2025)Jiang, Cohen, and Li]{jiang2025evolution}
Kaiqi Jiang, Jeremy Cohen, and Yuanzhi Li.
\newblock Understanding the evolution of the neural tangent kernel at the edge
  of stability.
\newblock In \emph{Advances in Neural Information Processing Systems},
  volume~38, 2025.
\newblock \doi{10.48550/arXiv.2507.12837}.
\newblock URL \url{https://arxiv.org/abs/2507.12837}.

\bibitem[Johnstone(2017)]{johnstone2017_GaussianEstimation}
Iain~M. Johnstone.
\newblock Gaussian estimation: {{Sequence}} and wavelet models.
\newblock Manuscript, 2017.

\bibitem[Karoui(2010)]{Karoui_spectrum_2010}
Noureddine~El Karoui.
\newblock {The spectrum of kernel random matrices}.
\newblock \emph{The Annals of Statistics}, 38\penalty0 (1):\penalty0 1 -- 50,
  2010.
\newblock \doi{10.1214/08-AOS648}.
\newblock URL \url{https://doi.org/10.1214/08-AOS648}.

\bibitem[Kobak et~al.(2020)Kobak, Lomond, and Sanchez]{kobak2020optimal}
Dmitry Kobak, Jonathan Lomond, and Benoit Sanchez.
\newblock The optimal ridge penalty for real-world high-dimensional data can be
  zero or negative due to the implicit ridge regularization.
\newblock \emph{Journal of Machine Learning Research}, 21\penalty0
  (169):\penalty0 1--16, 2020.

\bibitem[Kohler and Krzyżak(2001)]{Kohler2001NonparametricRE}
Michael Kohler and Adam Krzyżak.
\newblock Nonparametric regression estimation using penalized least squares.
\newblock \emph{IEEE Trans. Inf. Theory}, 47:\penalty0 3054--3059, 2001.

\bibitem[Kopitkov and Indelman(2020)]{kopitkov2020neural}
Dmitry Kopitkov and Vadim Indelman.
\newblock Neural spectrum alignment: Empirical study.
\newblock In \emph{Artificial Neural Networks and Machine Learning--ICANN 2020:
  29th International Conference on Artificial Neural Networks, Bratislava,
  Slovakia, September 15--18, 2020, Proceedings, Part II 29}, pages 168--179.
  Springer, 2020.

\bibitem[Kornblith et~al.(2019)Kornblith, Norouzi, Lee, and
  Hinton]{kornblith2019similarity}
Simon Kornblith, Mohammad Norouzi, Honglak Lee, and Geoffrey Hinton.
\newblock Similarity of neural network representations revisited.
\newblock In \emph{International conference on machine learning}, pages
  3519--3529. PMLR, 2019.

\bibitem[Lai et~al.(2023)Lai, Xu, Chen, and Lin]{lai2023generalization}
Jianfa Lai, Manyun Xu, Rui Chen, and Qian Lin.
\newblock Generalization ability of wide neural networks on $\mathbb{R}$.
\newblock \emph{arXiv preprint arXiv:2302.05933}, 2023.

\bibitem[Lecu{\'e} et~al.(2025)Lecu{\'e}, Li, and Shang]{lecue2025sharp}
Guillaume Lecu{\'e}, Zhifan Li, and Zong Shang.
\newblock Sharp convergence rates for spectral methods via the feature space
  decomposition method.
\newblock \emph{arXiv preprint arXiv:2512.14473}, 2025.
\newblock \doi{10.48550/arXiv.2512.14473}.
\newblock URL \url{https://arxiv.org/abs/2512.14473}.

\bibitem[Lee et~al.(2019)Lee, Xiao, Schoenholz, Bahri, Novak, Sohl-Dickstein,
  and Pennington]{lee2019wide}
Jaehoon Lee, Lechao Xiao, Samuel Schoenholz, Yasaman Bahri, Roman Novak, Jascha
  Sohl-Dickstein, and Jeffrey Pennington.
\newblock Wide neural networks of any depth evolve as linear models under
  gradient descent.
\newblock \emph{Advances in neural information processing systems}, 32, 2019.

\bibitem[Li et~al.(2021)Li, Nguyen, Hegde, and Wong]{li2021implicit}
Jiangyuan Li, Thanh Nguyen, Chinmay Hegde, and Ka~Wai Wong.
\newblock Implicit sparse regularization: The impact of depth and early
  stopping.
\newblock \emph{Advances in Neural Information Processing Systems},
  34:\penalty0 28298--28309, 2021.

\bibitem[Li and Lin(2024)]{li2024improving}
Yicheng Li and Qian Lin.
\newblock Improving adaptivity via over-parameterization in sequence models.
\newblock \emph{arXiv preprint arXiv:2409.00894}, 2024.

\bibitem[Li et~al.(2022)Li, Zhang, and Lin]{li2022saturation}
Yicheng Li, Haobo Zhang, and Qian Lin.
\newblock On the saturation effect of kernel ridge regression.
\newblock In \emph{The Eleventh International Conference on Learning
  Representations}, 2022.

\bibitem[Li et~al.(2023)Li, Zhang, and Lin]{10.1093/biomet/asad048}
Yicheng Li, Haobo Zhang, and Qian Lin.
\newblock {Kernel interpolation generalizes poorly}.
\newblock \emph{Biometrika}, page asad048, 08 2023.
\newblock ISSN 1464-3510.
\newblock \doi{10.1093/biomet/asad048}.
\newblock URL \url{https://doi.org/10.1093/biomet/asad048}.

\bibitem[Li et~al.(2024{\natexlab{a}})Li, Gan, Shi, and
  Lin]{li2024generalization}
Yicheng Li, Weiye Gan, Zuoqiang Shi, and Qian Lin.
\newblock Generalization error curves for analytic spectral algorithms under
  power-law decay.
\newblock \emph{arXiv preprint arXiv:2401.01599}, 2024{\natexlab{a}}.

\bibitem[Li et~al.(2024{\natexlab{b}})Li, Lin, et~al.]{li2023asymptotic}
Yicheng Li, Qian Lin, et~al.
\newblock On the asymptotic learning curves of kernel ridge regression under
  power-law decay.
\newblock \emph{Advances in Neural Information Processing Systems}, 36,
  2024{\natexlab{b}}.

\bibitem[Li et~al.(2024{\natexlab{c}})Li, Yu, Chen, and Lin]{li2024eigenvalue}
Yicheng Li, Zixiong Yu, Guhan Chen, and Qian Lin.
\newblock On the eigenvalue decay rates of a class of neural-network related
  kernel functions defined on general domains.
\newblock \emph{Journal of Machine Learning Research}, 25\penalty0
  (82):\penalty0 1--47, 2024{\natexlab{c}}.

\bibitem[Liang and Rakhlin(2020)]{Liang_Just_2019}
Tengyuan Liang and Alexander Rakhlin.
\newblock {Just interpolate: Kernel “Ridgeless” regression can generalize}.
\newblock \emph{The Annals of Statistics}, 48\penalty0 (3):\penalty0 1329 --
  1347, 2020.
\newblock \doi{10.1214/19-AOS1849}.
\newblock URL \url{https://doi.org/10.1214/19-AOS1849}.

\bibitem[Liang et~al.(2020)Liang, Rakhlin, and Zhai]{liang2020multiple}
Tengyuan Liang, Alexander Rakhlin, and Xiyu Zhai.
\newblock On the multiple descent of minimum-norm interpolants and restricted
  lower isometry of kernels.
\newblock In \emph{Conference on Learning Theory}, pages 2683--2711. PMLR,
  2020.

\bibitem[Lin and Cevher(2020)]{lin2020_OptimalConvergence}
Junhong Lin and Volkan Cevher.
\newblock Optimal convergence for distributed learning with stochastic gradient
  methods and spectral algorithms.
\newblock \emph{Journal of Machine Learning Research}, 21:\penalty0 147--1,
  2020.

\bibitem[Lin et~al.(2018)Lin, Rudi, Rosasco, and Cevher]{lin2018_OptimalRates}
Junhong Lin, Alessandro Rudi, L.~Rosasco, and V.~Cevher.
\newblock Optimal rates for spectral algorithms with least-squares regression
  over {{Hilbert}} spaces.
\newblock \emph{Applied and Computational Harmonic Analysis}, 48:\penalty0
  868--890, 2018.

\bibitem[Liu et~al.(2021)Liu, Liao, and Suykens]{Liu_kernel_2021}
Fanghui Liu, Zhenyu Liao, and Johan Suykens.
\newblock Kernel regression in high dimensions: Refined analysis beyond double
  descent.
\newblock In \emph{International Conference on Artificial Intelligence and
  Statistics}, pages 649--657. PMLR, 2021.

\bibitem[Lu et~al.(2023)Lu, Zhang, Li, Xu, and Lin]{lu2023optimal}
Weihao Lu, Haobo Zhang, Yicheng Li, Manyun Xu, and Qian Lin.
\newblock Optimal rate of kernel regression in large dimensions.
\newblock \emph{arXiv preprint arXiv:2309.04268}, 2023.

\bibitem[Maennel et~al.(2020)Maennel, Alabdulmohsin, Tolstikhin, Baldock,
  Bousquet, Gelly, and Keysers]{maennel2020neural}
Hartmut Maennel, Ibrahim~M Alabdulmohsin, Ilya~O Tolstikhin, Robert Baldock,
  Olivier Bousquet, Sylvain Gelly, and Daniel Keysers.
\newblock What do neural networks learn when trained with random labels?
\newblock \emph{Advances in Neural Information Processing Systems},
  33:\penalty0 19693--19704, 2020.

\bibitem[Mallinar et~al.(2022)Mallinar, Simon, Abedsoltan, Pandit, Belkin, and
  Nakkiran]{mallinar2022benign}
Neil Mallinar, James~B Simon, Amirhesam Abedsoltan, Parthe Pandit, Mikhail
  Belkin, and Preetum Nakkiran.
\newblock Benign, tempered, or catastrophic: A taxonomy of overfitting.
\newblock \emph{arXiv preprint arXiv:2207.06569}, 2022.

\bibitem[Mei and Montanari(2022)]{mei2022generalization2}
Song Mei and Andrea Montanari.
\newblock The generalization error of random features regression: Precise
  asymptotics and the double descent curve.
\newblock \emph{Communications on Pure and Applied Mathematics}, 75\penalty0
  (4):\penalty0 667--766, 2022.

\bibitem[Mei et~al.(2021)Mei, Misiakiewicz, and Montanari]{mei2021learning}
Song Mei, Theodor Misiakiewicz, and Andrea Montanari.
\newblock Learning with invariances in random features and kernel models.
\newblock In \emph{Conference on Learning Theory}, pages 3351--3418. PMLR,
  2021.

\bibitem[Mei et~al.(2022)Mei, Misiakiewicz, and
  Montanari]{mei2022generalization}
Song Mei, Theodor Misiakiewicz, and Andrea Montanari.
\newblock Generalization error of random feature and kernel methods:
  Hypercontractivity and kernel matrix concentration.
\newblock \emph{Applied and Computational Harmonic Analysis}, 59:\penalty0
  3--84, 2022.
\newblock ISSN 1063-5203.
\newblock \doi{https://doi.org/10.1016/j.acha.2021.12.003}.
\newblock URL
  \url{https://www.sciencedirect.com/science/article/pii/S1063520321001044}.

\bibitem[Misiakiewicz(2022)]{misiakiewicz_spectrum_2022}
Theodor Misiakiewicz.
\newblock Spectrum of inner-product kernel matrices in the polynomial regime
  and multiple descent phenomenon in kernel ridge regression.
\newblock \emph{arXiv preprint arXiv:2204.10425}, 2022.

\bibitem[Misiakiewicz and Saeed(2024)]{misiakiewicz2024non}
Theodor Misiakiewicz and Basil Saeed.
\newblock A non-asymptotic theory of kernel ridge regression: deterministic
  equivalents, test error, and gcv estimator.
\newblock \emph{arXiv preprint arXiv:2403.08938}, 2024.

\bibitem[Moniri et~al.(2023)Moniri, Lee, Hassani, and
  Dobriban]{moniri2023theory}
Behrad Moniri, Donghwan Lee, Hamed Hassani, and Edgar Dobriban.
\newblock A theory of non-linear feature learning with one gradient step in
  two-layer neural networks.
\newblock \emph{arXiv preprint arXiv:2310.07891}, 2023.

\bibitem[Montanari and Zhong(2022)]{montanari2022interpolation}
Andrea Montanari and Yiqiao Zhong.
\newblock The interpolation phase transition in neural networks: Memorization
  and generalization under lazy training.
\newblock \emph{The Annals of Statistics}, 50\penalty0 (5):\penalty0
  2816--2847, 2022.

\bibitem[Naveh and Ringel(2021)]{naveh2021self}
Gadi Naveh and Zohar Ringel.
\newblock A self consistent theory of gaussian processes captures feature
  learning effects in finite cnns.
\newblock \emph{Advances in Neural Information Processing Systems},
  34:\penalty0 21352--21364, 2021.

\bibitem[Nitanda and Suzuki(2020)]{nitanda2020optimal}
Atsushi Nitanda and Taiji Suzuki.
\newblock Optimal rates for averaged stochastic gradient descent under neural
  tangent kernel regime.
\newblock In \emph{International Conference on Learning Representations}, 2020.

\bibitem[Ortiz-Jim{\'e}nez et~al.(2021)Ortiz-Jim{\'e}nez, Moosavi-Dezfooli, and
  Frossard]{ortiz2021can}
Guillermo Ortiz-Jim{\'e}nez, Seyed-Mohsen Moosavi-Dezfooli, and Pascal
  Frossard.
\newblock What can linearized neural networks actually say about
  generalization?
\newblock \emph{Advances in Neural Information Processing Systems},
  34:\penalty0 8998--9010, 2021.

\bibitem[Oymak et~al.(2019)Oymak, Fabian, Li, and
  Soltanolkotabi]{oymak2019generalization}
Samet Oymak, Zalan Fabian, Mingchen Li, and Mahdi Soltanolkotabi.
\newblock Generalization guarantees for neural networks via harnessing the
  low-rank structure of the jacobian.
\newblock \emph{arXiv preprint arXiv:1906.05392}, 2019.

\bibitem[Pillaud-Vivien et~al.(2018)Pillaud-Vivien, Rudi, and
  Bach]{PillaudVivien2018StatisticalOO}
Loucas Pillaud-Vivien, Alessandro Rudi, and Francis Bach.
\newblock Statistical optimality of stochastic gradient descent on hard
  learning problems through multiple passes.
\newblock \emph{Advances in Neural Information Processing Systems}, 31, 2018.

\bibitem[Radhakrishnan et~al.(2022)Radhakrishnan, Beaglehole, Pandit, and
  Belkin]{radhakrishnan2022mechanism}
Adityanarayanan Radhakrishnan, Daniel Beaglehole, Parthe Pandit, and Mikhail
  Belkin.
\newblock Mechanism of feature learning in deep fully connected networks and
  kernel machines that recursively learn features.
\newblock \emph{arXiv preprint arXiv:2212.13881}, 2022.
\newblock \doi{10.48550/arXiv.2212.13881}.
\newblock URL \url{https://arxiv.org/abs/2212.13881}.

\bibitem[Radhakrishnan et~al.(2024)Radhakrishnan, Beaglehole, Pandit, and
  Belkin]{radhakrishnan2024mechanism}
Adityanarayanan Radhakrishnan, Daniel Beaglehole, Parthe Pandit, and Mikhail
  Belkin.
\newblock Mechanism for feature learning in neural networks and
  backpropagation-free machine learning models.
\newblock \emph{Science}, 383\penalty0 (6690):\penalty0 1461--1467, 2024.

\bibitem[Rahimi and Recht(2007)]{rahimi2007random}
Ali Rahimi and Benjamin Recht.
\newblock Random features for large-scale kernel machines.
\newblock \emph{Advances in neural information processing systems}, 20, 2007.

\bibitem[Rakhlin and Zhai(2019)]{rakhlin2019consistency}
Alexander Rakhlin and Xiyu Zhai.
\newblock Consistency of interpolation with laplace kernels is a
  high-dimensional phenomenon.
\newblock In \emph{Conference on Learning Theory}, pages 2595--2623. PMLR,
  2019.

\bibitem[Rosasco et~al.(2005)Rosasco, De~Vito, and
  Verri]{rosasco2005_SpectralMethods}
Lorenzo Rosasco, Ernesto De~Vito, and Alessandro Verri.
\newblock Spectral methods for regularization in learning theory.
\newblock \emph{DISI, Universita degli Studi di Genova, Italy, Technical Report
  DISI-TR-05-18}, 2005.

\bibitem[Sawano(2018)]{sawano2018theory}
Yoshihiro Sawano.
\newblock \emph{Theory of Besov spaces}, volume~56.
\newblock Springer, 2018.

\bibitem[Schmidt-Hieber(2020)]{schmidt2020nonparametric}
Johannes Schmidt-Hieber.
\newblock Nonparametric regression using deep neural networks with relu
  activation function.
\newblock \emph{The Annals of Statistics}, 48\penalty0 (4):\penalty0
  1875--1897, 2020.

\bibitem[Seleznova and Kutyniok(2022)]{seleznova2022finite}
Mariia Seleznova and Gitta Kutyniok.
\newblock Analyzing finite neural networks: Can we trust neural tangent kernel
  theory?
\newblock In \emph{Proceedings of the 2nd Mathematical and Scientific Machine
  Learning Conference}, volume 145 of \emph{Proceedings of Machine Learning
  Research}, pages 868--895, 2022.
\newblock URL \url{https://proceedings.mlr.press/v145/seleznova22a.html}.

\bibitem[Shan and Bordelon(2021)]{shan2021theory}
Haozhe Shan and Blake Bordelon.
\newblock A theory of neural tangent kernel alignment and its influence on
  training.
\newblock \emph{arXiv preprint arXiv:2105.14301}, 2021.

\bibitem[Smola et~al.(2000)Smola, Ov{\'a}ri, and
  Williamson]{smola2000_RegularizationDotproduct}
Alex Smola, Zolt{\'a}n Ov{\'a}ri, and Robert~C. Williamson.
\newblock Regularization with dot-product kernels.
\newblock \emph{Advances in neural information processing systems}, 13, 2000.

\bibitem[Steinwart and Christmann(2008)]{steinwart2008support}
Ingo Steinwart and Andreas Christmann.
\newblock \emph{Support vector machines}.
\newblock Springer Science \& Business Media, 2008.

\bibitem[Steinwart and Scovel(2012)]{steinwart2012_MercerTheorem}
Ingo Steinwart and C.~Scovel.
\newblock Mercer's theorem on general domains: {{On}} the interaction between
  measures, kernels, and {{RKHSs}}.
\newblock \emph{Constructive Approximation}, 35\penalty0 (3):\penalty0
  363--417, 2012.

\bibitem[Steinwart et~al.(2009)Steinwart, Hush, and
  Scovel]{steinwart2009_OptimalRates}
Ingo Steinwart, D.~Hush, and C.~Scovel.
\newblock Optimal rates for regularized least squares regression.
\newblock In \emph{{{COLT}}}, pages 79--93, 2009.

\bibitem[Suh and Cheng(2024)]{suh2024survey}
Namjoon Suh and Guang Cheng.
\newblock A survey on statistical theory of deep learning: Approximation,
  training dynamics, and generative models.
\newblock \emph{arXiv preprint arXiv:2401.07187}, 2024.

\bibitem[Suzuki(2018)]{suzuki2018adaptivity}
Taiji Suzuki.
\newblock Adaptivity of deep relu network for learning in besov and mixed
  smooth besov spaces: optimal rate and curse of dimensionality.
\newblock In \emph{International Conference on Learning Representations}, 2018.

\bibitem[Tartar(2007)]{tartar2007introduction}
Luc Tartar.
\newblock \emph{An introduction to Sobolev spaces and interpolation spaces},
  volume~3.
\newblock Springer Science \& Business Media, 2007.

\bibitem[Tirer et~al.(2022)Tirer, Bruna, and Giryes]{tirer2022kernel}
Tom Tirer, Joan Bruna, and Raja Giryes.
\newblock Kernel-based smoothness analysis of residual networks.
\newblock In \emph{Mathematical and Scientific Machine Learning}, pages
  921--954. PMLR, 2022.

\bibitem[Tsigler and Bartlett(2023)]{tsigler2023benign}
Alexander Tsigler and Peter~L Bartlett.
\newblock Benign overfitting in ridge regression.
\newblock \emph{Journal of Machine Learning Research}, 24\penalty0
  (123):\penalty0 1--76, 2023.

\bibitem[Vaskevicius et~al.(2019)Vaskevicius, Kanade, and
  Rebeschini]{vaskevicius2019implicit}
Tomas Vaskevicius, Varun Kanade, and Patrick Rebeschini.
\newblock Implicit regularization for optimal sparse recovery.
\newblock \emph{Advances in Neural Information Processing Systems}, 32, 2019.

\bibitem[Wainwright(2019)]{wainwright2019_HighdimensionalStatistics}
Martin~J. Wainwright.
\newblock \emph{High-Dimensional Statistics: {{A}} Non-Asymptotic Viewpoint}.
\newblock Cambridge {{Series}} in {{Statistical}} and {{Probabilistic
  Mathematics}}. {Cambridge University Press}, 2019.

\bibitem[Wang and Jing(2022)]{JMLR:v23:21-0570}
Wenjia Wang and Bing-Yi Jing.
\newblock Gaussian process regression: Optimality, robustness, and relationship
  with kernel ridge regression.
\newblock \emph{Journal of Machine Learning Research}, 23\penalty0
  (193):\penalty0 1--67, 2022.
\newblock URL \url{http://jmlr.org/papers/v23/21-0570.html}.

\bibitem[Wei et~al.(2019)Wei, Lee, Liu, and Ma]{wei2019regularization}
Colin Wei, Jason~D Lee, Qiang Liu, and Tengyu Ma.
\newblock Regularization matters: Generalization and optimization of neural
  nets vs their induced kernel.
\newblock \emph{Advances in Neural Information Processing Systems}, 32, 2019.

\bibitem[Woodworth et~al.(2020)Woodworth, Gunasekar, Lee, Moroshko, Savarese,
  Golan, Soudry, and Srebro]{woodworth2020kernel}
Blake Woodworth, Suriya Gunasekar, Jason~D Lee, Edward Moroshko, Pedro
  Savarese, Itay Golan, Daniel Soudry, and Nathan Srebro.
\newblock Kernel and rich regimes in overparametrized models.
\newblock In \emph{Conference on Learning Theory}, pages 3635--3673. PMLR,
  2020.

\bibitem[Xiao et~al.(2022)Xiao, Hu, Misiakiewicz, Lu, and
  Pennington]{xiao2022precise}
L~Xiao, H~Hu, T~Misiakiewicz, Y~Lu, and J~Pennington.
\newblock Precise learning curves and higher-order scaling limits for dot
  product kernel regression.
\newblock In \emph{Thirty-sixth Conference on Neural Information Processing
  Systems (NeurIPS)}, 2022.

\bibitem[Yaida(2020)]{yaida2020non}
Sho Yaida.
\newblock Non-gaussian processes and neural networks at finite widths.
\newblock In \emph{Mathematical and Scientific Machine Learning}, pages
  165--192. PMLR, 2020.

\bibitem[Yang and Hu(2020)]{yang2020feature}
Greg Yang and Edward~J Hu.
\newblock Feature learning in infinite-width neural networks.
\newblock \emph{arXiv preprint arXiv:2011.14522}, 2020.
\newblock \doi{10.48550/arXiv.2011.14522}.
\newblock URL \url{https://arxiv.org/abs/2011.14522}.

\bibitem[Zhang et~al.(2021)Zhang, Bengio, Hardt, Recht, and
  Vinyals]{zhang2021understanding}
Chiyuan Zhang, Samy Bengio, Moritz Hardt, Benjamin Recht, and Oriol Vinyals.
\newblock Understanding deep learning (still) requires rethinking
  generalization.
\newblock \emph{Communications of the ACM}, 64\penalty0 (3):\penalty0 107--115,
  2021.

\bibitem[Zhang et~al.(2023)Zhang, Li, Lu, and Lin]{zhang2023optimality_2}
Haobo Zhang, Yicheng Li, Weihao Lu, and Qian Lin.
\newblock On the optimality of misspecified kernel ridge regression.
\newblock In \emph{International Conference on Machine Learning}, pages
  41331--41353. PMLR, 2023.

\bibitem[Zhang et~al.(2024{\natexlab{a}})Zhang, Li, and
  Lin]{zhang2024optimality}
Haobo Zhang, Yicheng Li, and Qian Lin.
\newblock On the optimality of misspecified spectral algorithms.
\newblock \emph{Journal of Machine Learning Research}, 25\penalty0
  (188):\penalty0 1--50, 2024{\natexlab{a}}.

\bibitem[Zhang et~al.(2024{\natexlab{b}})Zhang, Li, Lu, and
  Lin]{zhang2024optimal}
Haobo Zhang, Yicheng Li, Weihao Lu, and Qian Lin.
\newblock Optimal rates of kernel ridge regression under source condition in
  large dimensions.
\newblock \emph{arXiv preprint arXiv:2401.01270}, 2024{\natexlab{b}}.

\bibitem[Zhang et~al.(2024{\natexlab{c}})Zhang, Lu, and Lin]{zhang2024phase}
Haobo Zhang, Weihao Lu, and Qian Lin.
\newblock {The phase diagram of kernel interpolation in large dimensions}.
\newblock \emph{Biometrika}, page asae057, 11 2024{\natexlab{c}}.
\newblock ISSN 1464-3510.
\newblock \doi{10.1093/biomet/asae057}.
\newblock URL \url{https://doi.org/10.1093/biomet/asae057}.

\bibitem[Zhao et~al.(2022)Zhao, Yang, and He]{zhao2022high}
Peng Zhao, Yun Yang, and Qiao-Chu He.
\newblock High-dimensional linear regression via implicit regularization.
\newblock \emph{Biometrika}, 109\penalty0 (4):\penalty0 1033--1046, 2022.

\end{thebibliography}
\end{document}